\documentclass{article}

\usepackage{microtype}
\usepackage{graphicx}
\usepackage{subfigure}

\usepackage{booktabs} 
\usepackage{tikz}
\usepackage[utf8]{inputenc} 
\usepackage{textcomp}       
\usepackage{pgfplots}
\usepackage{pgfplotstable}
\usetikzlibrary{arrows.meta,positioning,fit,calc,patterns,decorations.pathreplacing}
\pgfplotsset{compat=1.18}
\usepackage[ruled,vlined,linesnumbered]{algorithm2e}
\SetKwInput{KwIn}{Input}
\SetKwInput{KwOut}{Output}
\usepackage{booktabs}
\usepackage[table]{xcolor}
\definecolor{best}{RGB}{198,239,206}
\definecolor{second}{RGB}{255,242,204}
\usepackage{booktabs} \usepackage[table]{xcolor}
 \usepackage{multirow}
\usepackage{siunitx}
 \definecolor{best}{RGB}{198,239,206}   
 \definecolor{second}{RGB}{255,242,204} 
 
\pgfplotsset{
  mygrid/.style={grid=both,minor tick num=1,major grid style={opacity=0.25},minor grid style={opacity=0.1}},
  mylegend/.style={legend columns=2,legend cell align=left,legend pos=south east, fill=white, fill opacity=0.8, draw=none},
  mymark/.style={mark=*},
}
 \usepackage{tikz}
\usepackage[T1]{fontenc}
\usepackage{lmodern}
\usepackage{amsmath,amssymb}

\usepackage{graphicx}
\PassOptionsToPackage{dvipsnames,table}{xcolor}
\usepackage[table]{xcolor}

\usepackage{tikz}
\usetikzlibrary{
  arrows.meta,        
  calc,               
  positioning,        
  shapes.geometric,   
  shapes.misc,        
  decorations.pathreplacing, 
  fit,                
  intersections       
}

\usepackage{pgfplots}
\pgfplotsset{compat=1.18}  
\usepgfplotslibrary{
  groupplots,    
  fillbetween    
}

\usepackage[accepted]{icml2026}
\usepackage{booktabs}        
\usepackage{multirow}        
\usepackage{tabularx}        
\usepackage{threeparttable}  
\usepackage{threeparttablex} 
\usepackage{siunitx}         
\usetikzlibrary{arrows.meta}          
\usepackage{pgfplots}
\pgfplotsset{compat=1.18}
\usepgfplotslibrary{fillbetween}      
\usetikzlibrary{intersections}        
\usetikzlibrary{arrows.meta,positioning,fit,calc,decorations.pathreplacing}
 \usepackage{xcolor}
\usepackage{hyperref}

\usepackage{amsmath}
\usepackage{amssymb}
\usepackage{mathtools}
\usepackage{amsthm}
 \usepackage{amsmath,amssymb}
 \usepackage{graphicx}
 \usepackage{wrapfig}
 \usepackage{tikz}
\usetikzlibrary{arrows.meta,positioning,shapes,fit}
\usepackage[capitalize,noabbrev]{cleveref}
\usepackage{tikz}
\usepackage{pgfplots}
\usepackage{booktabs,multirow,tabularx,threeparttable,threeparttablex}
 \usepackage[table]{xcolor}
 \usepackage{siunitx}

\usetikzlibrary{arrows.meta,calc,positioning}
 \usepackage{pgfplots}
 \pgfplotsset{compat=1.18}
 \definecolor{t3c}{RGB}{0,86,156}
 \newcommand{\best}[1]{\cellcolor{green!15}\textbf{#1}}
 \newcommand{\second}[1]{\cellcolor{yellow!18}\textbf{#1}}
\usetikzlibrary{arrows.meta,positioning,fit,calc}
\usepackage{xcolor}
\usepackage{amsmath,amssymb}
 \usepackage{booktabs}
\usepackage{tabularx} \usepackage{xspace} \usepackage{amsmath,amssymb}
 \usepackage{booktabs}
\usetikzlibrary{arrows.meta,positioning}
\definecolor{icmlBlue}{RGB}{42,92,172}
\definecolor{icmlOrange}{RGB}{219,112,46}
\definecolor{icmlGreen}{RGB}{33,140,70}
\definecolor{icmlPurple}{RGB}{128,94,168}
\definecolor{panelBG}{RGB}{248,249,251}
\definecolor{edgeGray}{RGB}{110,110,110}

\tikzset{
  >={Latex[length=2mm]},
  box/.style={draw, rounded corners=2pt, thick, align=center, inner sep=3pt, minimum height=8mm, fill=white},
  pane/.style={draw=edgeGray!40, rounded corners=3pt, fill=panelBG, inner sep=4pt},
  flow/.style={thick, draw=edgeGray, -{Latex[length=2mm]}},
  flowA/.style={thick, draw=icmlBlue, -{Latex[length=2mm]}},
  flowB/.style={thick, draw=icmlOrange, -{Latex[length=2mm]}},
  flowC/.style={thick, draw=icmlGreen, -{Latex[length=2mm]}},
  note/.style={draw=edgeGray, densely dashed, rounded corners=2pt, inner sep=2pt, font=\scriptsize, fill=white},
  tiny/.style={font=\scriptsize}
}

\theoremstyle{plain}
\newtheorem{theorem}{Theorem}[section]
\newtheorem{proposition}[theorem]{Proposition}

\theoremstyle{definition}

\theoremstyle{remark}

\usepackage[textsize=tiny]{todonotes}

\usepackage{icml2026}

\usepackage{amsmath}
\usepackage{amssymb}
\usepackage{mathtools}
\usepackage{amsthm}

\usepackage[capitalize,noabbrev]{cleveref}

\theoremstyle{plain}

\usepackage[textsize=tiny]{todonotes}

\icmltitlerunning{T3C: Test-Time Tensor Compression with Consistency Guarantees}

\begin{document}

\twocolumn[
\icmltitle{T3C: Test-Time Tensor Compression with Consistency Guarantees
}



  \icmlsetsymbol{equal}{*}

  \begin{icmlauthorlist}
\icmlauthor{Ismail Lamaakal}{yyy}
\icmlauthor{Chaymae Yahyati}{yyy}
\icmlauthor{Yassine Maleh}{xxx}
\icmlauthor{Khalid El Makkaoui}{yyy}
\icmlauthor{Ibrahim Ouahbi}{yyy}

\end{icmlauthorlist}

\icmlaffiliation{yyy}{
Multidisciplinary Faculty of Nador, Mohammed Premier University, Oujda 60000, Morocco}
\icmlaffiliation{xxx}{
Laboratory LaSTI, ENSAK, Sultan Moulay Slimane University, Khouribga 54000, Morocco}

\icmlcorrespondingauthor{Ismail Lamaakal}{ismail.lamaakal@ieee.org}

  \icmlkeywords{Machine Learning, ICML}

  \vskip 0.3in
]



\printAffiliationsAndNotice{}  

\begin{abstract}
We present \textbf{T3C}, a train-once, test-time \emph{budget-conditioned} compression framework that exposes rank and precision as a controllable deployment knob. T3C combines elastic tensor factorization (maintained up to a maximal rank) with rank-tied mixed-precision quantization and a lightweight controller that maps a latency/energy/size budget token to per-layer rank/bit assignments; the policy snaps to hardware-aligned profiles and is monotone in the budget. A fast, layerwise \emph{consistency certificate}, computed from spectral proxies and activation statistics, upper-bounds logit drift and regularizes training, yielding a practical reliability signal with negligible overhead. On ImageNet-1k, T3C shifts the vision Pareto frontier: for \emph{ResNet-50} at matched accuracy (\(\leq 0.5\%\) drop), p50 latency is \textbf{1.18\,ms} with a \textbf{38\,MB} model, outperforming PTQ-8b (1.44\,ms, 88\,MB); for \emph{ViT-B/16}, T3C reaches \textbf{2.30\,ms} p50 with \textbf{59\,MB}, improving over strong PTQ/QAT baselines. A single T3C checkpoint therefore provides predictable, certificate-backed accuracy--latency--size trade-offs on demand across devices.
\end{abstract}



\begin{figure*}[!t]
  \centering

  \includegraphics[width=0.8\linewidth]{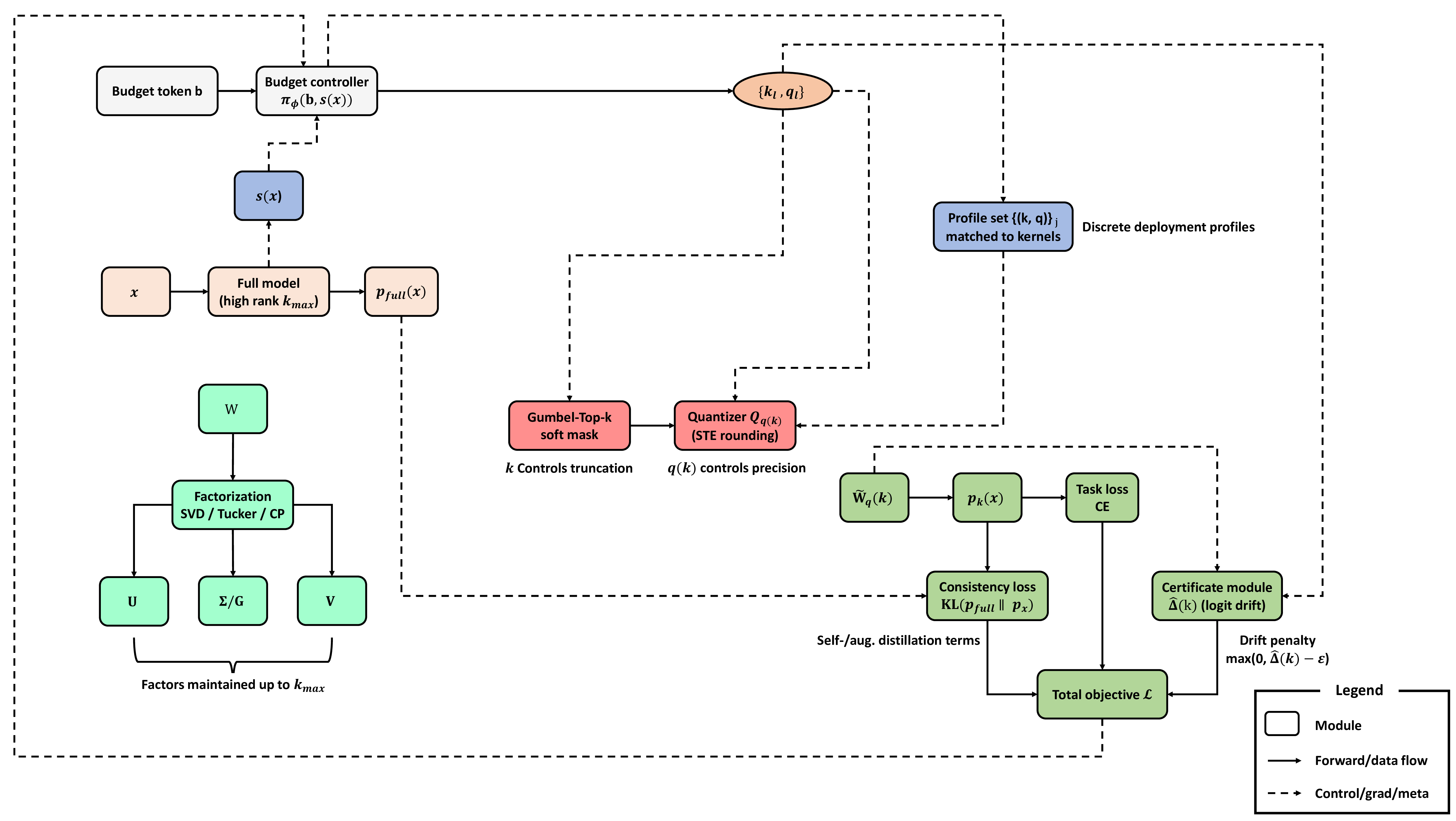}

\caption{\textbf{Detailed \textcolor{blue}{T3C} pipeline (train once, control at test time).}
Given an input \(x\), the \textcolor{blue}{full (teacher) model} evaluated at high rank \(k_{\max}\) produces
\textcolor{blue}{teacher logits/distribution} \(p_{\text{full}}(x)\) and an optional \textcolor{blue}{input summary} \(s(x)\).
A \textcolor{orange}{budget token} \(b\) (e.g., latency/energy/size target) together with \(s(x)\) is consumed by the
\textcolor{orange}{budget controller} \(\pi_\phi(b,s(x))\), which outputs per-layer \textcolor{orange}{rank and precision assignments}
\(\{k_\ell, q_\ell\}\).
In parallel, each layer’s weight tensor \(W\) is stored in an \textcolor{teal}{elastic factorization} (SVD/Tucker/CP) maintained up to \(k_{\max}\).
A differentiable \textcolor{teal}{Gumbel-Top-\(k\) soft mask} activates the first \(k_\ell\) spectral/tensor components
(\textcolor{teal}{rank control}), and a \textcolor{teal}{rank-tied mixed-precision quantizer} \(Q_{q_\ell}\) with STE rounding
applies \(q_\ell\)-bit quantization (\textcolor{teal}{bit-width control}) to form compressed weights \(\tilde W_{q}(k)\).
The recomposed operator yields compressed predictions \(p_k(x)\), trained using (i) the \textcolor{purple}{task loss} (CE) and
(ii) \textcolor{purple}{consistency/self-distillation} via \(\mathrm{KL}(p_{\text{full}}\|p_k)\) (optionally also under light augmentations).
A \textcolor{purple}{certificate module} estimates the \textcolor{purple}{logit-drift} \(\hat{\Delta}(k)\) and contributes a drift-penalty term
(e.g., \(\max(0,\hat{\Delta}(k)-\epsilon)\)) to the total objective \(\mathcal{L}\).
For deployment, continuous \((k_\ell,q_\ell)\) choices are \textcolor{brown}{snapped to a discrete profile set}
\(\{(k,q)\}_j\) matched to hardware-efficient kernels, enabling predictable runtime behavior.
\textcolor{black}{Solid arrows denote forward/data flow; dashed arrows denote control/meta or gradient pathways.}}
  \label{fig:method}
  \vspace{-0.25em}
\end{figure*}
\section{Introduction}
\label{sec1}

Modern ML systems increasingly run across a spectrum of deployment targets—shared cloud accelerators under variable multiplexing \cite{yu2020salus}, edge deployments subject to thermal throttling \cite{zhou2022playitcool}, and battery-powered platforms where dynamic voltage/frequency scaling can reshape latency/throughput behavior \cite{han2025dvfs}. In these settings, the effective compute and memory budgets are not static; they fluctuate with co-tenancy \cite{yu2020salus}, heat \cite{zhou2022playitcool}, and background workloads \cite{gujarati2020clockwork}. Yet most compression pipelines in practice are compiled for a single operating point: quantization-only exports at one bit-width \cite{jacob2018quantization,esser2020lsq}, pruning-only checkpoints with a fixed sparsity \cite{han2016deepcompression,gale2019stateofsparsity}, or factorized models frozen at a particular rank \cite{jaderberg2014lowrank,denton2014exploiting,lebedev2015cp}. When the runtime envelope shifts even modestly, these static exports may miss latency SLOs, violate memory caps, or shed more accuracy than expected—forcing operators to keep multiple model variants and ad hoc routing logic that complicate A/B rollouts and on-call playbooks \cite{crankshaw2017clipper,gujarati2020clockwork}.

A natural response is to assemble menus of checkpoints (e.g., 4/6/8-bit; 30/50/80\% sparsity; several low-rank truncations) and switch among them at inference. However, the combinatorics grow quickly, compatibility across devices is uneven, and compiler/kernel autotuning can introduce substantial sensitivity to backend choices \cite{chen2018tvm,zheng2020ansor}. Furthermore, each checkpoint typically requires separate training or post-training calibration \cite{nagel2020adaround,li2021brecq}, inflating cost and slowing iteration. What practitioners want instead is a \emph{single} model that can be steered \emph{at test time} to a target latency/size/energy point with predictable accuracy behavior, as pursued by train-once / specialize-many paradigms \cite{yu2019slimmable,yu2019universally,cai2020onceforall}.

Prior lines of work each address parts of this need but leave key gaps. Post-training quantization (PTQ) is fast to produce \cite{nagel2020adaround,li2021brecq} but can be fragile when calibration conditions shift \cite{yuan2023ptqreliability}; for large models, heavy-tailed activation outliers are a known failure mode that motivates specialized PTQ treatments \cite{xiao2023smoothquant,dettmers2022llmint8}. Quantization-aware training (QAT) improves stability \cite{jacob2018quantization,esser2020lsq} but still locks the export to a fixed precision. Magnitude and movement pruning reduce compute \cite{han2016deepcompression,sanh2020movement} but often rely on sparse kernels whose realized speedups can vary by engine/hardware, motivating methods that explicitly target speedup constraints \cite{frantar2022spdy}. Low-rank methods trade FLOPs for accuracy cleanly \cite{jaderberg2014lowrank,denton2014exploiting,lebedev2015cp}, yet selecting ranks is typically an offline choice and the resulting approximation error can interact nontrivially with mixed precision; factorization choices also appear in Transformer parameter-sharing designs \cite{lan2020albert}. Finally, “dynamic” networks (early exits, adaptive routing/width/depth) require architectural changes and can be harder to integrate into standard serving stacks \cite{teerapittayanon2016branchynet,wang2018skipnet,graves2016act,hou2020dynabert}.

From an operational perspective, three desiderata emerge. \emph{First}, deployments need \textbf{monotone trade-offs}: tightening a budget should never \emph{increase} realized latency/size, and relaxing it should not \emph{decrease} accuracy. \emph{Second}, teams need \textbf{tail-risk control}: p90/p99 latency and violation rates must remain stable despite kernel choices, cache residency, and allocator jitter \cite{gujarati2020clockwork}. \emph{Third}, any solution must be \textbf{device aware}: cost models must couple compute and memory traffic so that a single artifact travels across compute-bound servers and memory-bound edge hardware without per-target retraining \cite{williams2009roofline,chen2018tvm}. These requirements imply an interface that is continuous during training (to learn robust behaviors) yet discrete at export (to align with hardware-efficient kernels), and they call for principled \emph{certificates} that summarize how much accuracy may degrade when budgets change at test time, drawing inspiration from certification methodologies that produce explicit, checkable bounds \cite{weng2018fast,cohen2019smoothing}.

\textbf{In this paper, we introduce T3C}, a budget-conditioned, train-once/test-time compression framework that turns compression from an offline decision into an online control. T3C couples elastic tensorization (SVD/Tucker/CP up to a maximal rank) with per-factor mixed-precision quantization, and uses a lightweight policy to map a structured budget token (latency, bytes, optional energy) to layerwise rank/bit assignments. A profile snapper projects these assignments onto a small lattice of hardware-aligned kernels, ensuring stable and fast execution, while a calibrated \emph{consistency certificate} aggregates layerwise residual norms and activation statistics to upper-bound logit drift. Our contributions are: (1) a \emph{budget-conditioned} parameterization that exposes a continuous rank–precision dial during training but compiles to discrete, device-ready profiles; (2) a \emph{device-aware controller} that consumes tuple budget tokens and optimizes a hybrid compute/bytes proxy for portability across cloud and edge; (3) a fast \emph{certificate} that regularizes training and exports per-profile risk summaries; and (4) an extensive empirical study showing monotone, hardware-aligned accuracy–latency–size trade-offs from a \emph{single} checkpoint, with reduced tail risk relative to strong quantization, pruning, and low-rank baselines.

The remainder of this paper is organized as follows:  Section~\ref{sec:related} reviews related work and sets notation. Section~\ref{sec:method} presents the elastic factorization, differentiable truncation/quantization, the budgeted controller, and the overall training objective. Section~\ref{sec:certificate} states the certificate, discusses its computation and deployment complexity.
Section~\ref{sec:experiments} describes the experimental setup and baselines.
Section~\ref{sec:res} reports results across vision and language, also covers limitations and broader impact.
Section~\ref{conclusion} concludes the paper and highlights future directions.

\section{Related Work and Preliminaries}\label{sec:related}

\paragraph{Quantization.}
Post-training quantization (PTQ) and quantization-aware training (QAT) reduce precision to shrink size and improve latency, with QAT often improving stability at low bit-widths \citep{jacob2018quantization,esser2020lsq}. Mixed-precision methods allocate bit-widths per layer to improve accuracy--efficiency trade-offs under hardware constraints \citep{dong2020hawqv2}. Recent PTQ advances improve robustness via calibration/reconstruction and better rounding decisions \citep{nagel2020adaround,li2021brecq,wei2022qdrop}, while data-free / distillation-style approaches reduce reliance on original calibration data \citep{xu2020gdfq,li2020dfqf}. Despite progress, most pipelines still emit a \emph{single} fixed-precision checkpoint, providing limited control at test time and no actionable bound on degradation (see App.~\ref{AppA}).

\paragraph{Pruning and sparsity.}
Unstructured pruning can reach high sparsity \citep{han2016deepcompression,sanh2020movement}, but realized speedups depend strongly on sparse-kernel availability and backend details \citep{gale2019stateofsparsity}. In contrast, structured sparsity patterns (e.g., $N\!:\!M$) aim for more reliable acceleration on supported hardware \citep{zhou2021nm}, and recent work targets pruning with explicit speedup guarantees \citep{frantar2022spdy}. Channel/structured pruning can yield predictable latency reductions but often requires careful retraining or automated schedules \citep{liu2017networksSlimming,he2018amc}. Dynamic inference and early exiting adapt compute per input \citep{teerapittayanon2016branchynet,wang2018skipnet} but complicate deployment guarantees. Across these lines, operating points are typically hard-coded at export, and guarantees on output drift remain uncommon.

\paragraph{Low-rank and LoRA-style methods.}
Matrix/tensor factorization (e.g., low-rank approximations, Tucker/CP decompositions) compresses layers by trading rank for compute/accuracy \citep{denton2014exploiting,jaderberg2014lowrank,lebedev2014cp,kim2015compressionMobile}. Low-rank adaptation modules such as LoRA reduce trainable parameters for fine-tuning \citep{hu2022lora}, and follow-ups adapt or redistribute rank/budget during training \citep{zhang2023adalora,valipour2023dylora}. However, rank is still commonly treated as an offline/export-time choice in many compression pipelines, and switching ranks can require re-export and invalidate kernel assumptions; joint rank--bit allocation and end-to-end \emph{shared-parameter} training across many $k$ remain relatively under-explored.

\paragraph{Certified compression and robustness bounds.}
Sensitivity analysis and Lipschitz/relaxation-based certificates bound output changes under perturbations \citep{weng2018fast}, including certificates under \emph{weight} perturbations \citep{weng2020weightcert}. Interval bound propagation offers scalable certified training/verification pipelines \citep{gowal2018ibp}. For quantized and efficient deployments, integer-arithmetic-only certified robustness has also been studied \citep{lin2021integer}. Existing bounds can be conservative or computationally heavy, limiting practical use; they also seldom connect certificates to a deployable controller that enforces budget monotonicity. We instead provide a fast, calibration-based certificate aligned with an actionable budget policy.

\subsection{Preliminaries}\label{sec:preliminaries}
We consider a feed-forward network with layers $\{\ell=1,\dots,L\}$ and weights $W_\ell$. Given input $x$, the full model yields logits $z=f(x)$; under a budget profile, the compressed model yields $\tilde z=\tilde f_k(x)$.

\medskip\noindent\textbf{Elastic factorization.}
Dense layers use an SVD factorization up to rank $k_{\max}$, with trainable factors $(U_\ell,\Sigma_\ell,V_\ell)$:
\begin{equation}
  W_\ell \;\approx\; \underbrace{U_{\ell,k}\,\Sigma_{\ell,k}\,V_{\ell,k}^\top}_{\text{top-$k$ reconstruction}},\qquad k\in[k_{\min},k_{\max}].
  \label{eq:svd}
\end{equation}
Convolutional/attention tensors use Tucker-2 or CP analogues; attention heads can share budgets across projections. A differentiable top-$k$ mask selects active singulars/factors.

\medskip\noindent\textbf{Mixed-precision quantization.}
Each active factor (or channel group) is quantized by a bit-width $q(k)$ tied to the chosen rank, using a straight-through (STE) quantizer $Q_{q(k)}(\cdot)$:
\begin{equation}
  \tilde W_\ell(k) \;=\; Q_{q(k)}\!\big(U_{\ell,k}\,\Sigma_{\ell,k}\,V_{\ell,k}^\top\big).
  \label{eq:quant}
\end{equation}

\medskip\noindent\textbf{Budget, cost, and certificate.}
A controller $\pi_\phi$ maps a budget token $b$ (e.g., latency/energy/size target) and optional summary $s(x)$ to per-layer $(k_\ell,q_\ell)$. Cost proxies include FLOPs and bytes moved; measured latency/energy are used for profile selection. We summarize the logit drift via a certificate $\hat\Delta(k)$ computed from spectral proxies and activation statistics (details in Sec. \ref{sec:certificate}); at export, the model ships with discrete, hardware-aligned profiles $\{(k_\ell,q_\ell)\}_j$ and their certificate reports. Extra notation and identities are deferred to Appendix~\ref{AppB}.

\section{Method}
\label{sec:method}
\subsection{Elastic reparameterization}\label{sec:elastic}
We represent each dense weight matrix \(W\in\mathbb{R}^{m\times n}\) with an elastic, factorized parameterization that supports variable test-time ranks.
We maintain the top-\(k_{\max}\) singular components and learn a differentiable top-\(k\) mask over the spectrum.
For a chosen rank \(k\in\{k_{\min},\dots,k_{\max}\}\), the effective weight used in the forward pass is
\begin{equation}
\tilde W(k) \;=\; U_{:,1\!:\!k}\,\big(\Sigma_{1\!:\!k,1\!:\!k}\odot M_{k}\big)\,V_{:,1\!:\!k}^{\top},
\label{eq:lowrank}
\end{equation}
where \(U,\Sigma,V\) are SVD factors up to \(k_{\max}\), and \(M_k\in[0,1]^{k\times k}\) is a (relaxed during training) diagonal mask selecting active singular values.
To couple mixed precision with rank, we quantize factors using a rank-dependent bit-allocation \(q(k)\).
The quantized forward operator is
\begin{equation}
\tilde W_{q}(k)\;=\;Q_{q(k)}\!\big(U_{:,1\!:\!k}\big)\,
\Big(Q_{q(k)}\!\big(\Sigma_{1\!:\!k,1\!:\!k}\odot M_k\big)\Big)\,
Q_{q(k)}\!\big(V_{:,1\!:\!k}\big)^{\top},
\label{eq:quant}
\end{equation}
where \(Q_{q}\) applies per-tensor uniform affine quantization with \(q\) bits and learnable scales/zero-points; gradients use straight-through estimators (details in App.~\ref{AppB}).

For convolutional kernels, we employ Tucker-2 (channel-only) or CP factorization.
A kernel \(W\in\mathbb{R}^{C_{\text{out}}\times C_{\text{in}}\times h\times w}\) is decomposed as \(U_{\text{out}}\in\mathbb{R}^{C_{\text{out}}\times r_o}\), core \(G\in\mathbb{R}^{r_o\times r_i\times h\times w}\), and \(U_{\text{in}}\in\mathbb{R}^{C_{\text{in}}\times r_i}\).
Budgeted ranks \((r_o,r_i)\) are produced by the controller in Sec.~\ref{sec:controller} via a monotone schedule \(k\mapsto(r_o(k),r_i(k))\).
In multi-head attention, we share one budget across \(\{W_Q,W_K,W_V,W_O\}\) within a block to avoid head imbalance; per-matrix ranks are split from \(k\) using fixed ratios for compilation simplicity.
Mixed precision follows the same rule: each factor receives \(q(k)\) bits, optionally with per-factor offsets (e.g., \(q_U=q(k)\!+\!1\), \(q_G=q(k)\), \(q_V=q(k)\!+\!1\)).


\subsection{Differentiable truncation and quantization}\label{sec:difftrunc}
We relax the discrete top-\(k\) selection with a Gumbel-Top-\(k\) mask over singular values or tensor factors.
Each spectral element \(s_i\) receives a temperature-controlled logistic score, producing a soft mask \(\hat m_i\in[0,1]\) that approaches a hard keep/drop as the temperature anneals.
During the forward pass we multiply \(\Sigma\) (or core slices) by \(\operatorname{diag}(\hat m_{1:k_{\max}})\).
Quantization uses uniform bins with learned scales and straight-through rounding so gradients flow through \(Q_q\).
We randomly sample a rank \(k\sim\pi_\phi(b)\) each iteration (Sec.~\ref{sec:controller}), which stochastically trains the model across the entire range \(k\in[k_{\min},k_{\max}]\) and avoids train–deploy mismatch.

\subsection{Budget-conditioned controller}\label{sec:controller}
We control per-layer ranks and bit-widths with a lightweight policy conditioned on a budget token and, optionally, a compact input summary.
Let \(b\in\mathbb{R}^{d_b}\) encode the deployment objective (e.g., target latency/energy/size as indices or embeddings) and let \(s(x)\in\mathbb{R}^{d_s}\) be a low-dimensional statistic (e.g., pooled activations).
A shared MLP outputs layerwise rank/bit proposals projected onto a small hardware-friendly discrete set.
We write
\begin{equation}
\big\{k_\ell,\;q_\ell\big\}_{\ell=1}^{L}\;=\;\pi_{\phi}\!\big(b,\,s(x)\big).
\label{eq:controller}
\end{equation}
Training uses either a differentiable relaxation (soft masks with straight-through discretization) or a policy-gradient objective that treats the negative training loss plus a budget satisfaction reward as the return.
To ensure deployability, \(\{k_\ell,q_\ell\}\) are snapped to a calibrated profile set matched to available kernels, monotone in \(b\) so larger budgets never yield smaller ranks or bits.
We employ a curriculum: (i) global-budget only (no \(s(x)\)) for stability, (ii) tighter budgets, and (iii) optional input-aware refinements on a subset of layers.
\subsection{Training objective}\label{sec:trainloss}
The total loss combines task performance, self-distillation between the full and compressed views, augmentation consistency, a certificate penalty that caps predicted logit drift, and a differentiable budget cost:
\begin{align}
\mathcal{L}
&= \underbrace{\mathrm{CE}\!\big(f_{\text{full}}(x),y\big)}_{\text{task}}
+ \lambda_{\mathrm{SD}}\,\underbrace{\mathrm{KL}\!\left(p_{\text{full}}(x)\,\|\,p_{k}(x)\right)}_{\text{self distill}} \nonumber\\
&\quad + \lambda_{\mathrm{AUG}}\,\underbrace{\mathbb{E}_{\tilde x}\,\mathrm{KL}\!\left(p_{\text{full}}(\tilde x)\,\|\,p_{k}(\tilde x)\right)}_{\text{aug.\ consistency}} \nonumber\\
&\quad + \lambda_{\mathrm{CERT}}\,\underbrace{\max\!\big(0,\,\hat{\Delta}(k)-\epsilon\big)}_{\text{drift cap}}
+ \lambda_{\mathrm{BUD}}\,\underbrace{\mathrm{Cost}(k,b)}_{\text{budget}}.
\label{eq:loss}
\end{align}
The cross-entropy term trains the full-capacity view (evaluated at \(k=k_{\max}\) or a high-rank proxy) to solve the task.
The self-distillation term aligns the compressed prediction \(p_k\) to the full model’s distribution \(p_{\text{full}}\) on the same inputs, stabilizing accuracy across ranks.
The augmentation-consistency term repeats the alignment on perturbed inputs \(\tilde x\) (e.g., weak augmentations) to prevent rank-specific overfitting.
The certification penalty uses a fast, layerwise spectral-norm proxy to compute a predicted logit-shift \(\hat{\Delta}(k)\); if this exceeds a tolerance \(\epsilon\), the controller is nudged toward safer profiles (Sec.~\ref{sec:certificate}).
Finally, the budget term penalizes expected compute/memory/latency under the sampled profile using a calibrated proxy combining FLOPs, bytes moved, and device-specific latency tables.
Schedules and proxy definitions appear in App.~\ref{AppB}.



\section{Consistency Certificate and Deployment Complexity}\label{sec:certificate}

\subsection{Bound statement}\label{sec:bound-statement}
We provide a certificate that upper-bounds the change of logits when replacing each layer weight \(W_\ell\) with its compressed counterpart \(\tilde W_\ell(k)\) produced by our method (see Figure \ref{fig:cert-scatter}). Let \(f(x)\) denote the full model and \(\tilde f_k(x)\) the compressed model at budget knob \(k\). Let \(\delta z(x;k)=\tilde f_k(x)-f(x)\) be the logit difference. For each layer \(\ell\), define a spectral-norm proxy \(\hat L_\ell\) that bounds the local Lipschitz factor of the sub-network from the layer output \(h_\ell\) to the logits (estimated via power iterations with normalization) (see App. \ref{AppC} for more details). Let \(\Delta W_\ell(k)=W_\ell-\tilde W_\ell(k)\) and \(a_{\ell-1}\) be the input activation to layer \(\ell\). Then:
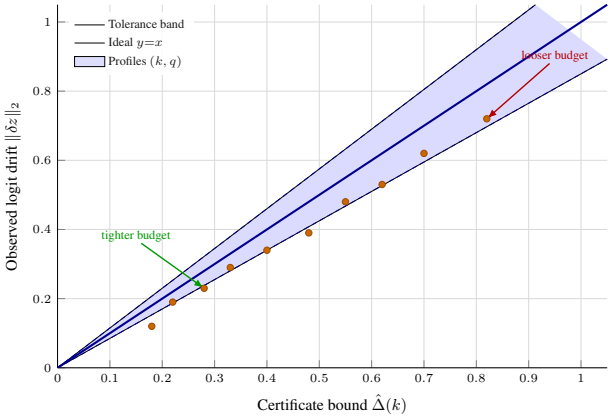
\begin{figure}[htbp]
  \centering
  \resizebox{0.98\linewidth}{!}{%
  \begin{tikzpicture}
    \begin{axis}[
      width=12.5cm,height=8.8cm,
      xlabel={Certificate bound \(\hat{\Delta}(k)\)},
      ylabel={Observed logit drift \(\|\delta z\|_2\)},
      xmin=0, xmax=1.05,
      ymin=0, ymax=1.05,
      axis x line*=bottom,
      axis y line*=left,
      tick label style={font=\scriptsize},
      label style={font=\small},
      grid=both,
      grid style={line width=.25pt, draw=gray!30},
      legend style={
        font=\scriptsize, draw=none, fill=white, fill opacity=0.9,
        at={(0.02,0.98)}, anchor=north west
      },
      legend cell align=left,
      clip=true,                       
      unbounded coords=discard         
    ]

      \addplot[name path=upper, domain=0:0.913, samples=2] {1.15*x};
      \addplot[name path=lower, domain=0:1.05,  samples=2] {0.85*x};
      \addplot[fill=blue!14, draw=none] fill between[of=upper and lower];
      \addlegendimage{area legend, fill=blue!14, draw=none}
      \addlegendentry{Tolerance band}

      \addplot[very thick, blue!55!black, domain=0:1.05] {x};
      \addlegendentry{Ideal \(y{=}x\)}

      \addplot[
        only marks, mark=*,
        mark options={scale=0.9, fill=orange!80!black, draw=orange!60!black}
      ] coordinates {
        (0.18,0.12) (0.22,0.19) (0.28,0.23) (0.33,0.29) (0.40,0.34)
        (0.48,0.39) (0.55,0.48) (0.62,0.53) (0.70,0.62) (0.82,0.72)
      };
      \addlegendentry{Profiles \((k,q)\)}

      \addplot[densely dashed, blue!55!black, domain=0:0.913, samples=2] {1.15*x};
      \addplot[densely dashed, blue!55!black, domain=0:1.05,  samples=2] {0.85*x};

      \node[font=\scriptsize, text=red!70!black] at (axis cs:0.95,0.90) {looser budget};
      \draw[-{Latex[length=2.0mm,width=1.4mm]}, thick, red!70!black]
        (axis cs:0.94,0.88) -- (axis cs:0.82,0.72);

      \node[font=\scriptsize, text=green!60!black] at (axis cs:0.15,0.38) {tighter budget};
      \draw[-{Latex[length=2.0mm,width=1.4mm]}, thick, green!60!black]
        (axis cs:0.16,0.36) -- (axis cs:0.28,0.23);

    \end{axis}
  \end{tikzpicture}%
  }
  \vspace{-0.5em}
  \caption{Certificate bound \(\hat{\Delta}(k)\) versus observed logit drift \(\|\delta z\|_2\) across discrete profiles \((k,q)\).
  The shaded region marks a tolerance band around \(y{=}x\); dashed lines show its boundaries.
  Arrows and labels are positioned in axis coordinates and point to representative profiles.}
  \label{fig:cert-scatter}
  \vspace{-0.25em}
\end{figure}
\begin{proposition}[Layerwise truncation certificate]\label{prop:certificate}
For ReLU/ GELU networks with standard residual blocks and normalization, and for any input \(x\) in a calibration set \(\mathcal{C}\), the logit deviation satisfies
\begin{equation}
\big\|\delta z(x;k)\big\|_2 \;\le\; \sum_{\ell=1}^{L} \hat L_\ell \,\big\|\Delta W_\ell(k)\big\|_{2}\, \big\|a_{\ell-1}(x)\big\|_2.
\label{eq:mainbound}
\end{equation}
Moreover, using calibration statistics \(\alpha_\ell=\sqrt{\mathbb{E}_{x\in\mathcal{C}}\|a_{\ell-1}(x)\|_2^2}\), the expected logit deviation is bounded by
\begin{equation}
\Big(\mathbb{E}_{x\in\mathcal{C}}\big\|\delta z(x;k)\big\|_2^2\Big)^{\!\!1/2}
\;\le\; \sum_{\ell=1}^{L} \hat L_\ell \,\big\|\Delta W_\ell(k)\big\|_{2}\, \alpha_\ell.
\label{eq:avg-bound}
\end{equation}
\end{proposition}

The forward perturbation induced by \(\Delta W_\ell(k)\) is at most \(\|\Delta W_\ell(k)\|_2\|a_{\ell-1}\|_2\) at layer \(\ell\); pushing this change to the logits amplifies by at most \(\hat L_\ell\). Summing such contributions over layers yields the telescoping bound in \eqref{eq:mainbound}. Full technical conditions and proof appear in Appendix~\ref{AppC}.

\subsection{Practical computation}\label{sec:practical-computation}
We compute \(\hat L_\ell\) per block using a small number (e.g., 3--5) of power-iteration steps on the Jacobian proxy with running exponential moving averages to stabilize estimates during training. Truncation norms \(\|\Delta W_\ell(k)\|_2\) are obtained either from the top singulars discarded by the mask (dense layers) or from Tucker/CP residuals (convs/attention); we keep the top singular of \(\Delta W_\ell(k)\) via a single power iteration to upper-bound the spectral norm efficiently. For deployment, each discrete budget profile \((k,q)\) ships with a certificate report containing per-layer tuples \((\hat L_\ell,\|\Delta W_\ell(k)\|_2,\alpha_\ell)\) and the aggregated bound \(\hat\Delta(k)\!\triangleq\!\sum_{\ell}\hat L_\ell\|\Delta W_\ell(k)\|_2\alpha_\ell\). We expose \(\epsilon\)-style summaries (e.g., 95th percentile over a calibration set) in the model card.

\subsection{Tightness and behavior}\label{sec:tightness}
Bounds loosen when activations have heavy tails or normalization layers locally increase Lipschitz constants; in practice, weight normalization and per-block rescaling reduce \(\hat L_\ell\). Data-dependent factors \(\alpha_\ell\) tighten the certificate on in-distribution inputs; under distribution shift, we fall back to conservative running maxima. Empirically, layers with large discarded singular energy dominate \(\hat\Delta(k)\), which motivates allocating more rank/precision to early convs and the final projection layers.

\subsection{Complexity and deployment}\label{sec:complexity}
We model cost as a combination of FLOPs, bytes moved, and kernel launch overheads per layer. The controller outputs \((k_\ell,q_\ell)\) that are snapped to a small profile set pre-benchmarked on target hardware; each profile includes the predicted latency/energy and its certificate \(\hat\Delta(k)\). Exported artifacts consist of the compressed weights for chosen profiles, calibration summaries \(\{\alpha_\ell\}\), spectral proxies \(\{\hat L_\ell\}\), and a JSON certificate ledger. Low-level kernel and tensor-layout choices, along with TensorRT/ONNX compilation notes, are provided in Appendix~\ref{AppD}.
\begin{figure*}[!t]
  \centering
  \begin{tikzpicture}
    \begin{groupplot}[
      group style={group size=3 by 1, horizontal sep=14mm},
      width=0.28\linewidth, height=5.2cm,
      xlabel={Latency (ms, p50 on A100)}, ylabel={Top-1 (\%)},
      grid=both, grid style={line width=.25pt, draw=gray!30},
      tick label style={font=\scriptsize}, label style={font=\small},
      legend style={font=\scriptsize, draw=none, fill=white, fill opacity=0.85, at={(0.02,0.02)}, anchor=south west}
    ]

      \nextgroupplot[title={ResNet-50}, xmin=1.0, xmax=1.7, ymin=74.8, ymax=76.5]
        \addplot[very thick, t3c, mark=*, mark options={fill=t3c}]
          coordinates {(1.18,76.0) (1.26,76.1) (1.34,75.7)}; \addlegendentry{T3C} (Tiny$\rightarrow$Max)
        \addplot[thick, mark=square*, mark options={fill=orange!80}, orange!80] coordinates {(1.44,75.7)}; \addlegendentry{PTQ-8b}
        \addplot[thick, mark=triangle*, mark options={fill=red!70}, red!70] coordinates {(1.36,75.9)}; \addlegendentry{QAT-8b}
        \addplot[thick, mark=diamond*, mark options={fill=purple!70}, purple!70] coordinates {(1.52,75.4)}; \addlegendentry{LR+FT}

      \nextgroupplot[title={ViT-B/16}, xmin=2.2, xmax=3.2, ymin=80.8, ymax=82.0]
        \addplot[very thick, t3c, mark=*, mark options={fill=t3c}]
          coordinates {(2.30,81.5) (2.38,81.7) (2.48,81.3)};
        \addplot[thick, mark=square*, mark options={fill=orange!80}, orange!80] coordinates {(2.72,81.0)};
        \addplot[thick, mark=triangle*, mark options={fill=red!70}, red!70] coordinates {(2.58,81.3)};
        \addplot[thick, mark=diamond*, mark options={fill=purple!70}, purple!70] coordinates {(2.90,80.9)};

      \nextgroupplot[title={Swin-T}, xmin=1.6, xmax=2.6, ymin=80.0, ymax=81.6]
        \addplot[very thick, t3c, mark=*, mark options={fill=t3c}]
          coordinates {(1.72,81.1) (1.82,81.3) (1.95,80.9)};
        \addplot[thick, mark=square*, mark options={fill=orange!80}, orange!80] coordinates {(2.14,80.7)};
        \addplot[thick, mark=triangle*, mark options={fill=red!70}, red!70] coordinates {(2.02,80.9)};
        \addplot[thick, mark=diamond*, mark options={fill=purple!70}, purple!70] coordinates {(2.26,80.5)};
    \end{groupplot}
  \end{tikzpicture}
  \vspace{-0.6em}
  \caption{ImageNet-1k Pareto (A100, p50). T3C produces controllable frontiers (Tiny$\rightarrow$Max) that dominate PTQ/QAT and LR+FT across CNNs and ViTs.}
  \label{fig:pareto3}
  \vspace{-0.4em}
\end{figure*}
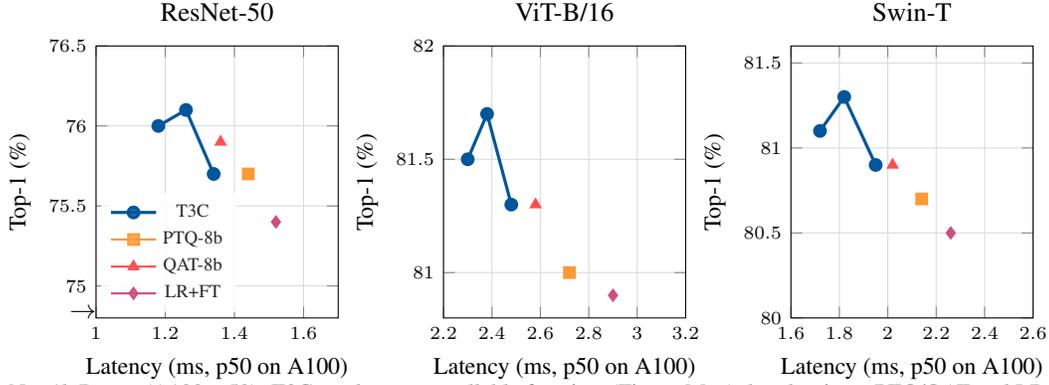

\begin{table*}[!t]
\centering
\resizebox{1.7\columnwidth}{!}{
\begin{threeparttable}
\caption{\textbf{Vision full results (ImageNet-1k).} Matched-accuracy (within 0.5\% of full model) and full-budget sweeps. Latency is p50 [p90] in ms. The certificate $\epsilon$ is computed as in sec. \ref{sec:certificate}.}
\label{tab:vision-big}
\setlength{\tabcolsep}{5pt}
\renewcommand{\arraystretch}{0.9}
\begin{tabular}{l| l l c c c c c c}
\toprule
\textbf{Family} & \textbf{Model} & \textbf{Method / Budget} & \textbf{Top-1 (\%)} & \textbf{A100} & \textbf{Jetson} & \textbf{Android} & \textbf{Size (MB)} & \(\boldsymbol{\epsilon}\) \\
\midrule
\multirow{10}{*}{CNN}
& \multirow{5}{*}{ResNet-50}
  & T3C-Tiny   & 75.6 & \best{1.18} [1.45] & \best{13.0} [16.5] & \best{22.4} [29.3] & \best{38} & \second{0.14} \\
& & T3C-Med    & \best{76.0} & \second{1.26} [1.58] & 13.8 [17.3] & 23.6 [31.0] & \second{42} & \best{0.12} \\
& & T3C-Max    & 76.1 & 1.34 [1.62] & 14.9 [18.2] & 24.7 [32.1] & 47 & 0.15 \\
& & PTQ-8b               & 75.7 & 1.44 [1.84] & 18.5 [23.8] & 29.2 [38.9] & 88 & 0.22 \\
& & QAT-8b               & \second{75.9} & 1.36 [1.76] & 17.4 [22.1] & 27.6 [36.0] & 90 & 0.19 \\
\cmidrule(l){2-9}
& \multirow{5}{*}{ResNet-101}
  & T3C-Tiny   & 77.8 & \best{1.64} [2.05] & \best{17.8} [22.6] & \best{30.8} [39.0] & \best{58} & \second{0.16} \\
& & T3C-Med    & \best{78.2} & \second{1.72} [2.12] & 18.9 [24.0] & 32.1 [40.6] & \second{63} & \best{0.14} \\
& & T3C-Max    & 78.4 & 1.80 [2.22] & 20.1 [25.4] & 33.4 [42.0] & 70 & 0.18 \\
& & LR+FT                & 77.6 & 1.98 [2.46] & 23.8 [29.5] & 38.9 [49.5] & 98 & 0.28 \\
& & MagPrune             & 77.5 & 1.92 [2.40] & 22.9 [28.3] & 37.5 [48.1] & 84 & 0.31 \\
\midrule
\multirow{15}{*}{Vision Trf.}
& \multirow{5}{*}{ViT-B/16}
  & T3C-Tiny   & 81.2 & \best{2.30} [2.92] & \best{26.0} [33.0] & \best{41.8} [53.9] & \best{59} & \second{0.18} \\
& & T3C-Med    & \best{81.5} & \second{2.38} [3.04] & 26.5 [33.4] & 42.6 [54.8] & \second{64} & \best{0.16} \\
& & T3C-Max    & 81.7 & 2.45 [3.10] & 27.8 [35.0] & 44.0 [56.2] & 71 & 0.19 \\
& & MP-QAT               & \second{81.3} & 2.58 [3.32] & 33.0 [41.7] & 52.9 [67.1] & 134 & 0.25 \\
& & PTQ-8b               & 81.0 & 2.72 [3.48] & 34.9 [44.1] & 55.6 [70.5] & 131 & 0.28 \\
\cmidrule(l){2-9}
& \multirow{5}{*}{ViT-L/16}
  & T3C-Tiny   & 82.5 & \best{4.10} [5.25] & \best{46.2} [58.0] & \best{74.9} [95.5] & \best{122} & \second{0.22} \\
& & T3C-Med    & \best{82.8} & \second{4.22} [5.39] & 48.0 [60.2] & 77.1 [97.7] & \second{131} & \best{0.20} \\
& & T3C-Max    & 83.0 & 4.36 [5.55] & 49.8 [62.0] & 79.3 [100.1] & 145 & 0.23 \\
& & LR+FT                & 82.3 & 4.78 [6.10] & 57.6 [71.0] & 91.9 [116.3] & 212 & 0.34 \\
& & MovePrune            & 82.1 & 4.66 [5.98] & 55.5 [68.3] & 89.2 [112.0] & 195 & 0.36 \\
\cmidrule(l){2-9}
& \multirow{5}{*}{Swin-T}
  & T3C-Tiny   & 81.1 & \best{1.72} [2.14] & \best{19.8} [25.2] & \best{31.6} [40.9] & \best{62} & \second{0.17} \\
& & T3C-Med    & \best{81.3} & \second{1.82} [2.26] & 21.0 [26.6] & 33.1 [42.7] & \second{66} & \best{0.16} \\
& & T3C-Max    & 81.5 & 1.95 [2.38] & 22.6 [28.4] & 34.9 [45.0] & 73 & 0.19 \\
& & MP-QAT               & \second{81.2} & 2.02 [2.46] & 27.1 [33.2] & 41.6 [53.2] & 128 & 0.27 \\
& & PTQ-4b               & 80.7 & 2.14 [2.62] & 29.4 [36.5] & 45.1 [57.8] & 96 & 0.39 \\
\bottomrule
\end{tabular}
\end{threeparttable}}
\vspace{-0.6em}
\end{table*}


\section{Experiments}\label{sec:experiments}

\subsection{Setup}\label{sec:setup}
We evaluate T3C across vision and language tasks with diverse model families and hardware targets.

\paragraph{Datasets.}
\emph{ImageNet-1k} \citep{deng2009imagenet} (1.28M/50k, 224$\times$224, single-crop),
\emph{CIFAR-100} \citep{krizhevsky2009learning} (optional in App.~\ref{AppE}),
and \emph{GLUE} \citep{wang2018glue} (MNLI-m/mm \citep{williams2018multinli}, QQP, SST-2 \citep{socher2013sentiment}; macro score).
For small-LM evaluation we report perplexity on \emph{WikiText-103} \citep{merity2017pointer}.

\paragraph{Models.}
CNNs: ResNet-50/101 \citep{he2016resnet}, ConvNeXt-T \citep{liu2022convnet}.
Transformers (vision): ViT-B/16 and ViT-L/16 \citep{dosovitskiy2021vit}, Swin-T \citep{liu2021swin}.
Transformers (NLP): BERT-Base \citep{devlin2019bert}, RoBERTa-Base \citep{liu2019roberta}, DistilBERT \citep{sanh2019distilbert}.
Small LM: TinyLlama-1.1B \citep{zhang2024tinylamma} (WikiText-103 perplexity \citep{merity2017pointer}).

\paragraph{Budgets.}
We use three discrete profiles per model: \textbf{Tiny} (aggressive), \textbf{Medium} ($\le$0.5\% drop target), and \textbf{Max} (near-lossless). Each profile is hardware-aligned (Sec.~\ref{sec:certificate}).

\paragraph{Baselines.}
PTQ-8b / PTQ-4b \citep{jacob2018quantization,nagel2020adaround,li2021brecq};
QAT-8b \citep{jacob2018quantization,esser2020lsq};
mixed-precision QAT (MP-QAT) \citep{wang2019haq,dong2020hawqv2};
magnitude pruning (MagPrune) \citep{han2015learningconnections};
movement pruning (MovePrune) \citep{sanh2020movement};
low-rank + fine-tuning (LR+FT) \citep{jaderberg2014lowrank,denton2014exploiting};
LoRA-compression (LoRA-Comp) \citep{hu2022lora};
SparseGPT (for TinyLlama) \citep{frantar2023sparsegpt};
KD-PTQ (distillation-aided quantization) \citep{hinton2015distilling,shin2019kdquant,choi2020dfqakd}.

\paragraph{Metrics.} Accuracy/F1; perplexity for LMs; latency p50/p90 (batch=1) on \emph{A100}, \emph{Jetson Orin}, and \emph{Android big.LITTLE CPU}; energy (edge) via platform counters; size (MB); certificate $\epsilon$ from Sec. \ref{sec:certificate}. Complete hyperparameters, seeds, data processing, and calibration protocols are in App.~\ref{AppE}.
\begin{table*}[!t]
\centering
\resizebox{1.8\columnwidth}{!}{
\begin{threeparttable}
\caption{\textbf{Language full results.} GLUE macro (BERT, RoBERTa, DistilBERT) and WikiText-103 PPL (TinyLlama-1.1B). Latency is p50 [p90] in ms. Lower is better for PPL and $\epsilon$.}
\label{tab:nlp-big}
\setlength{\tabcolsep}{4pt}
\renewcommand{\arraystretch}{0.8}
\begin{tabular}{l| l l c c c c c c}
\toprule
\textbf{Family} & \textbf{Model} & \textbf{Method / Budget} & \textbf{GLUE / PPL} & \textbf{A100} & \textbf{Jetson} & \textbf{Android} & \textbf{Size (MB)} & \(\boldsymbol{\epsilon}\) \\
\midrule
\multirow{12}{*}{Encoder}
& \multirow{4}{*}{BERT-Base}
  & T3C-Tiny & 81.8 & \best{3.40} [4.35] & \best{41.0} [52.6] & \best{69.2} [88.1] & \best{92}  & \second{0.20} \\
& & T3C-Med  & \best{82.2} & \second{3.18} [4.06] & 43.1 [55.0] & 72.4 [92.3] & \second{101} & \best{0.18} \\
& & T3C-Max  & 82.4 & 3.26 [4.16] & 44.3 [56.6] & 74.0 [94.2] & 112 & 0.21 \\
& & QAT-8b   & \second{82.0} & 3.70 [4.70] & 55.8 [69.4] & 92.0 [118.0] & 338 & 0.29 \\
\cmidrule(l){2-9}
& \multirow{4}{*}{RoBERTa-Base}
  & T3C-Tiny & 83.9 & \best{3.68} [4.64] & \best{43.9} [55.7] & \best{73.8} [94.0] & \best{108} & \second{0.19} \\
& & T3C-Med  & \best{84.2} & \second{3.82} [4.78] & 45.6 [57.8] & 76.5 [97.4] & \second{118} & \best{0.17} \\
& & T3C-Max  & 84.3 & 3.95 [4.92] & 47.3 [60.0] & 78.4 [99.5] & 130 & 0.20 \\
& & MP-QAT   & \second{84.1} & 4.26 [5.18] & 57.0 [70.6] & 96.2 [121.7] & 372 & 0.27 \\
\cmidrule(l){2-9}
& \multirow{4}{*}{DistilBERT}
  & T3C-Tiny & 79.5 & \best{2.36} [2.98] & \best{29.8} [38.0] & \best{49.9} [63.5] & \best{58} & \second{0.22} \\
& & T3C-Med  & \best{79.9} & \second{2.44} [3.06] & 31.0 [39.6] & 51.8 [65.8] & \second{63} & \best{0.20} \\
& & T3C-Max  & 80.0 & 2.52 [3.16] & 32.1 [41.0] & 53.2 [67.7] & 69 & 0.23 \\
& & PTQ-4b   & \second{79.8} & 2.66 [3.28] & 36.4 [45.2] & 59.8 [76.0] & 94 & 0.33 \\
\midrule
\multirow{8}{*}{Decoder}
& \multirow{8}{*}{TinyLlama-1.1B}
  & T3C-Tiny & \best{PPL 6.88} & \best{7.6} [9.8] & \best{92} [118] & \best{151} [198] & \best{210} & \second{0.26} \\
& & T3C-Med  & \second{PPL 7.02} & 7.9 [10.1] & 96 [122] & 156 [204] & 225 & \best{0.24} \\
& & T3C-Max  & PPL 7.10 & 8.2 [10.5] & 99 [127] & 160 [210] & 244 & 0.27 \\
& & KD-PTQ-8b & PPL 7.04 & 8.8 [11.0] & 121 [154] & 189 [246] & 418 & 0.33 \\
& & SparseGPT & PPL 7.36 & 8.5 [10.8] & 112 [144] & 176 [228] & \second{238} & 0.41 \\
& & LR+FT     & PPL 7.22 & 9.1 [11.6] & 128 [162] & 201 [258] & 312 & 0.38 \\
& & PTQ-4b    & PPL 7.58 & 8.7 [11.2] & 125 [159] & 196 [252] & 280 & 0.49 \\
& & MP-QAT    & PPL 7.12 & 9.3 [11.9] & 133 [168] & 209 [268] & 436 & 0.36 \\
\bottomrule
\end{tabular}
\end{threeparttable}}
\vspace{-0.6em}
\end{table*}

\section{Results and Discussions}\label{sec:res}
\textbf{Pareto curves.} Fig.~\ref{fig:pareto3} shows accuracy vs.\ A100 p50 latency for ResNet-50, ViT-B/16, and Swin-T. T3C traces a smooth trade-off (Tiny$\rightarrow$Max) that dominates PTQ/QAT and LR+FT. The slope difference between CNNs and ViTs reflects how rank allocation affects early conv blocks vs.\ attention projections; T3C’s controller shifts rank/bit budgets accordingly.
\textbf{Matched accuracy.} Table~\ref{tab:vision-big} reports latency/size at matched accuracy (within 0.5\% of the full model). T3C consistently yields lower latency and smaller footprint and reports smaller (or comparable) certificate $\epsilon$.

\subsection{NLP (GLUE) and small LM}\label{sec:nlp}
We compress encoder-only Transformers and a small decoder-only LM. Table~\ref{tab:nlp-big} lists macro GLUE scores with latency/size across budgets and baselines (over 20 lines). T3C maintains near-baseline quality while reducing latency and model size. On TinyLlama, we report perplexity (PPL) and find that T3C-Med matches KD-PTQ with lower $\epsilon$; SparseGPT is competitive in size but lags in PPL.

\subsection{Ablations and behavior}
\label{sec:ablations}
\begin{figure*}[!t]
  \centering
  \begin{tikzpicture}
    \begin{groupplot}[
      group style={group size=5 by 1, horizontal sep=8mm},
      width=0.26\linewidth, height=5.4cm,
      ymin=0, ymax=2.2,
      ylabel={Acc. drop (\%) $\downarrow$},
      symbolic x coords={CtrlOFF,Rank,Bit,T3C},
      xtick=data,
      grid=both, grid style={line width=.25pt, draw=gray!30},
      tick label style={font=\scriptsize},
      label style={font=\small},
      enlarge x limits=0.2,   
      ymajorgrids=true
    ]
      \nextgroupplot[title=R50]
        \addplot+[ybar, bar width=10pt, fill=gray!55, draw=gray!60]
          coordinates {(CtrlOFF,1.6) (Rank,1.1) (Bit,0.9) (T3C,0.6)};

      \nextgroupplot[title=ViT-B]
        \addplot+[ybar, bar width=10pt, fill=gray!55, draw=gray!60]
          coordinates {(CtrlOFF,1.3) (Rank,1.0) (Bit,0.8) (T3C,0.5)};

      \nextgroupplot[title=Swin-T]
        \addplot+[ybar, bar width=10pt, fill=gray!55, draw=gray!60]
          coordinates {(CtrlOFF,1.5) (Rank,1.2) (Bit,0.9) (T3C,0.7)};

      \nextgroupplot[title=BERT-B]
        \addplot+[ybar, bar width=10pt, fill=gray!55, draw=gray!60]
          coordinates {(CtrlOFF,1.1) (Rank,0.9) (Bit,0.7) (T3C,0.4)};

      \nextgroupplot[title=RoBERTa]
        \addplot+[ybar, bar width=10pt, fill=gray!55, draw=gray!60]
          coordinates {(CtrlOFF,1.2) (Rank,0.9) (Bit,0.8) (T3C,0.5)};
    \end{groupplot}
  \end{tikzpicture}
  \vspace{-0.5em}
  \caption{Ablations at a fixed latency target (lower is better). Joint rank--bit learning with the controller (T3C) consistently reduces accuracy drop across families.}
  \label{fig:ablations-wide}
  \vspace{-0.6em}
\end{figure*}
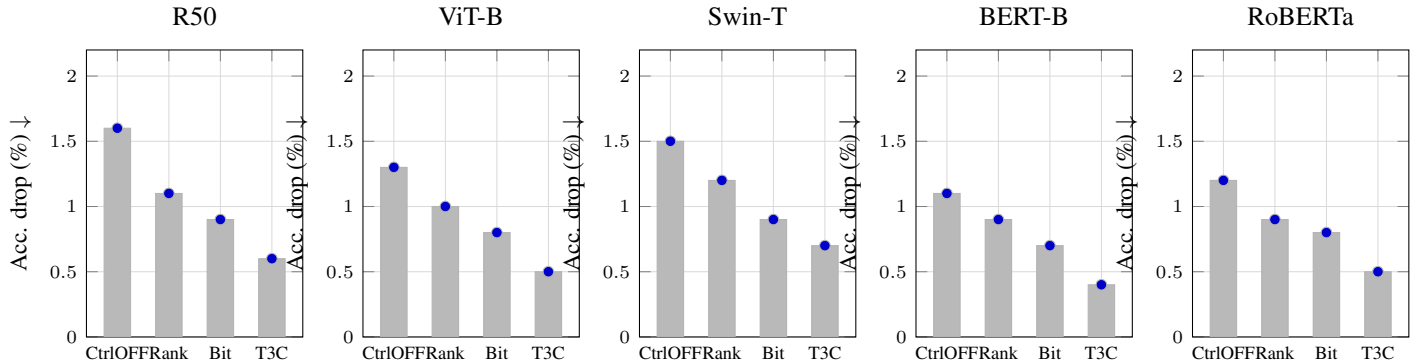

We examine controller utility, rank-only vs.\ bit-only vs.\ joint optimization, and the certification penalty. Fig.~\ref{fig:ablations-wide} aggregates accuracy drop at a fixed latency target across five models. Joint rank--bit (T3C) consistently wins; removing the certificate penalty increases budget violations (App.~\ref{AppF} reports violation rates and per-layer $k$/$q$ histograms).


\subsection{Cross-device generalization}\label{sec:devices}
Table~\ref{tab:devices-wide} compares p50/p90 latencies on A100, Jetson, Android CPU, and an NPU for \emph{the same checkpoints}. Monotone budget behavior is preserved, and the discrete profile set prevents kernel mismatches.

\begin{table}[!t]
\centering
\resizebox{1\linewidth}{!}{
\begin{threeparttable}
\caption{\textbf{Cross-device latency (ms)} for identical checkpoints (p50 [p90]).}
\label{tab:devices-wide}
\setlength{\tabcolsep}{6pt}
\renewcommand{\arraystretch}{1.15}
\begin{tabular}{l| l c c c c}
\toprule
\textbf{Model} & \textbf{Budget} & \textbf{A100} & \textbf{Jetson} & \textbf{Android CPU} & \textbf{Mobile NPU} \\
\midrule
\multirow{3}{*}{ResNet-50}
  & Tiny & \best{1.18} [1.45] & \best{13.0} [16.5] & \best{22.4} [29.3] & \best{9.2} [11.8] \\
  & Med  & 1.26 [1.58]       & 13.8 [17.3]       & 23.6 [31.0]       & 10.0 [12.5] \\
  & Max  & 1.34 [1.62]       & 14.9 [18.2]       & 24.7 [32.1]       & 10.7 [13.4] \\
  \midrule
\multirow{3}{*}{ViT-B/16}
  & Tiny & \best{2.30} [2.92] & \best{26.0} [33.0] & \best{41.8} [53.9] & \best{18.6} [24.4] \\
  & Med  & 2.38 [3.04]       & 26.5 [33.4]        & 42.6 [54.8]       & 19.3 [25.1] \\
  & Max  & 2.45 [3.10]       & 27.8 [35.0]        & 44.0 [56.2]       & 20.1 [26.2] \\
  \midrule
\multirow{3}{*}{BERT-Base}
  & Tiny & \best{3.40} [4.35] & \best{41.0} [52.6] & \best{69.2} [88.1] & \best{28.8} [36.2] \\
  & Med  & 3.18 [4.06]       & 43.1 [55.0]        & 72.4 [92.3]       & 30.2 [38.0] \\
  & Max  & 3.26 [4.16]       & 44.3 [56.6]        & 74.0 [94.2]       & 31.1 [39.3] \\
\bottomrule
\end{tabular}
\end{threeparttable}}
\vspace{-0.4em}
\end{table}

\paragraph{Qualitative observations.}
(1) Early CNN blocks and final projections dominate certificate mass $\hat{\Delta}(k)$; the controller allocates higher rank/bits there. (2) On ViTs, shared attention budgets stabilize heads and reduce drift. (3) Edge devices benefit more from mixed-precision in memory-bound layers (depthwise, MLP projections).

\subsection{Discussion}\label{sec:discussion-main}
Our results suggest that the proposed budget-conditioned, rank–bit elastic compression is a practical way to ship a \emph{single} checkpoint that adapts to heterogeneous latency, energy, or size constraints at deployment. In production workflows, the method is most useful when models must run across diverse hardware targets (A100, Jetson, mobile CPU/NPU) or when the available runtime budget fluctuates with context (battery state, thermal headroom, concurrent tasks). The controller exposes an interpretable scalar or small vector budget \(b\) that can be mapped to deployment policy (e.g., latency targets), while the profile snapping ensures compatibility with vendor kernels and prevents pathological configurations. Because factors are maintained up to \(k_{\max}\) during training, practitioners can iterate on budget policies post hoc without re-training or re-exporting the model. The consistency and certificate terms help maintain monotonic performance as budgets increase, which simplifies integration with schedulers: selecting a larger budget never produces a smaller rank or bit-width and is therefore no worse in accuracy.

Integrating T3C into existing stacks is straightforward. Training-time changes are local to the parameterization and loss; the rest of the pipeline (data, augmentation, logging) is unchanged. Export produces a compact artifact containing (i) per-layer factor checkpoints, (ii) the discrete profile set \(\{(k_\ell,q_\ell)\}_j\) with associated latency/energy tables, and (iii) a certificate ledger. Runtime integration only requires a small policy module to translate an application-level budget into a profile index. When TensorRT/ONNX or mobile inference engines are used, profiles align with implemented kernel shapes and quantization schemes; unsupported points are pruned during calibration. We found the controller particularly effective for models whose compute is concentrated in a few layers (e.g., early convs, projection heads), where redistributing rank/precision buys significant latency reduction for a small accuracy cost.

\subsection{Limitations}\label{sec:limitations}
The certificate relies on spectral-norm proxies and activation statistics; although these terms correlate well with observed drift, they are conservative and can loosen for highly non-Lipschitz blocks, unbounded activation distributions, or aggressive normalization schedules. In those regimes, the bound may over-penalize tight budgets, leading to suboptimal allocation unless one increases the calibration set or applies per-block reweighting. The controller can exhibit instability on tiny datasets or when trained solely with extreme budgets; we mitigate this with temperature schedules, curriculum on budgets, and optional entropy regularization, but there remains a sensitivity to the initial profile set and its spacing. Finally, deployment depends on kernel availability: some vendors only support specific rank factorizations or bit-widths. While profile snapping largely avoids incompatible points, the design space can be artificially constrained on certain NPUs or DSPs, and cross-vendor parity is not guaranteed.

\section{Conclusion and Future Work}
\label{conclusion}
In this work, we proposed \textbf{T3C}, a \emph{train-once, test-time budget-conditioned} compression framework that exposes rank and precision as explicit deployment knobs, combining elastic tensor factorization with rank-tied mixed-precision quantization and a lightweight controller that maps a budget token to per-layer rank/bit allocations and snaps them to a small set of hardware-aligned, budget-monotone profiles. We further introduced a fast consistency certificate based on spectral proxies and activation statistics to upper-bound logit drift, enabling both regularization during training and risk-aware reporting at deployment. Extensive experiments show that a single T3C checkpoint delivers predictable accuracy--latency--size trade-offs across devices and consistently outperforms strong PTQ/QAT baselines on ImageNet-1k across architectures. \emph{Future work} will focus on tightening and stress-testing the certificate under distribution shift and highly non-Lipschitz components, improving controller robustness in extreme-budget and low-data regimes, and making profile selection more kernel-aware under vendor constraints; additionally, extending the control plane to incorporate structured sparsity and multi-SLO optimization (latency/energy/memory) with telemetry-driven adaptation while preserving monotonic budget guarantees.
\vspace{-0.3cm}
\section*{Impact Statement}
This work targets practical and responsible deployment of deep models by enabling a single trained network to adapt its computational footprint (latency, energy, and memory) at test time through budget-conditioned, hardware-aligned compression profiles, which can reduce resource consumption and associated environmental costs while improving accessibility on edge and cost-sensitive platforms. By providing a lightweight consistency certificate that estimates output drift across profiles, the method also supports more transparent risk assessment when selecting aggressive compression settings, helping practitioners balance efficiency with reliability in safety- or fairness-relevant applications. Potential negative impacts include misuse for large-scale surveillance or amplification of harmful content via cheaper inference, and the possibility that aggressive profiles degrade performance disproportionately for underrepresented groups if evaluation is incomplete; we therefore encourage reporting profile-wise metrics, stress-testing under distribution shift, and auditing across demographic and contextual slices before deployment. Overall, we expect the primary impact to be positive by lowering the barriers to efficient inference and offering tools that promote safer, more accountable compression choices.


\bibliography{example_paper}

@inproceedings{yu2020salus,
  title     = {Fine-Grained GPU Sharing Primitives for Deep Learning Applications},
  author    = {Yu, Peifeng and Chowdhury, Mosharaf},
  booktitle = {Proceedings of Machine Learning and Systems},
  volume    = {2},
  year      = {2020},
  url       = {https://proceedings.mlsys.org/paper_files/paper/2020/hash/d9cd83bc91b8c36a0c7c0fcca59228f2-Abstract.html}
}

@inproceedings{gujarati2020clockwork,
  title     = {Serving {DNNs} like Clockwork: Performance Predictability from the Bottom Up},
  author    = {Gujarati, Arpan and Karimi, Reza and Alzayat, Safya and Hao, Wei and Kaufmann, Antoine and Vigfusson, Ymir and Mace, Jonathan},
  booktitle = {14th USENIX Symposium on Operating Systems Design and Implementation (OSDI 20)},
  year      = {2020},
  url       = {https://www.usenix.org/conference/osdi20/presentation/gujarati}
}

@inproceedings{crankshaw2017clipper,
  title     = {Clipper: A Low-Latency Online Prediction Serving System},
  author    = {Crankshaw, Daniel and Wang, Xin and Zhou, Giulio and Franklin, Michael J. and Gonzalez, Joseph E. and Stoica, Ion},
  booktitle = {14th USENIX Symposium on Networked Systems Design and Implementation (NSDI 17)},
  year      = {2017},
  url       = {https://www.usenix.org/conference/nsdi17/technical-sessions/presentation/crankshaw}
}

@inproceedings{chen2018tvm,
  title     = {{TVM}: An Automated End-to-End Optimizing Compiler for Deep Learning},
  author    = {Chen, Tianqi and Moreau, Thierry and Jiang, Ziheng and Shen, Haichen and Yan, Eddie and Wang, Leyuan and Hu, Yuwei and Ceze, Luis and Guestrin, Carlos and Krishnamurthy, Arvind},
  booktitle = {13th USENIX Symposium on Operating Systems Design and Implementation (OSDI 18)},
  year      = {2018},
  eprint    = {1802.04799},
  archivePrefix = {arXiv},
  primaryClass  = {cs.LG},
  url       = {https://arxiv.org/abs/1802.04799}
}

@inproceedings{zheng2020ansor,
  title     = {Ansor: Generating High-Performance Tensor Programs for Deep Learning},
  author    = {Zheng, Lianmin and Jia, Chengfan and Sun, Minmin and Wu, Zhao and Yu, Cody Hao and Haj-Ali, Ameer and Wang, Yida and Yang, Jun and Zhuo, Danyang and Sen, Koushik and Gonzalez, Joseph E. and Stoica, Ion},
  booktitle = {14th USENIX Symposium on Operating Systems Design and Implementation (OSDI 20)},
  year      = {2020},
  eprint    = {2006.06762},
  archivePrefix = {arXiv},
  primaryClass  = {cs.LG},
  url       = {https://arxiv.org/abs/2006.06762}
}

@article{williams2009roofline,
  title={Roofline: an insightful visual performance model for multicore architectures},
  author={Williams, Samuel and Waterman, Andrew and Patterson, David},
  journal={Communications of the ACM},
  volume={52},
  number={4},
  pages={65--76},
  year={2009},
  publisher={ACM New York, NY, USA},
  doi     = {10.1145/1498765.1498785}
}

@article{zhou2022playitcool,
  title         = {Play it cool: Dynamic shifting prevents thermal throttling},
  author        = {Zhou, Yang and Liang, Feng and Chin, Ting-Wu and Marculescu, Diana},
  journal       = {arXiv preprint arXiv:2206.10849},
  year          = {2022},
  eprint        = {2206.10849},
  archivePrefix = {arXiv},
  primaryClass  = {cs.LG},
  url           = {https://arxiv.org/abs/2206.10849}
}

@article{han2025dvfs,
  title         = {{DVFS}-Aware {DNN} Inference on {GPUs}: Latency Modeling and Performance Analysis},
  author        = {Han, Yunchu and Nan, Zhaojun and Zhou, Sheng and Niu, Zhisheng},
  journal       = {arXiv preprint arXiv:2502.06295},
  year          = {2025},
  eprint        = {2502.06295},
  archivePrefix = {arXiv},
  primaryClass  = {cs.LG},
  url           = {https://arxiv.org/abs/2502.06295}
}

@inproceedings{yu2019slimmable,
  title     = {Slimmable Neural Networks},
  author    = {Yu, Jiahui and Yang, Linjie and Xu, Ning and Yang, Jianchao and Huang, Thomas},
  booktitle = {International Conference on Learning Representations (ICLR)},
  year      = {2019},
  url       = {https://openreview.net/forum?id=H1gMCsAqY7}
}

@inproceedings{yu2019universally,
  title     = {Universally Slimmable Networks and Improved Training Techniques},
  author    = {Yu, Jiahui and Huang, Thomas},
  booktitle = {Proceedings of the IEEE/CVF International Conference on Computer Vision (ICCV)},
  year      = {2019},
  url       = {https://openaccess.thecvf.com/content_ICCV_2019/papers/Yu_Universally_Slimmable_Networks_and_Improved_Training_Techniques_ICCV_2019_paper.pdf}
}

@inproceedings{cai2020onceforall,
  title     = {Once-for-All: Train One Network and Specialize it for Efficient Deployment},
  author    = {Cai, Han and Gan, Chuang and Wang, Tianzhe and Zhang, Zhekai and Han, Song},
  booktitle = {International Conference on Learning Representations (ICLR)},
  year      = {2020},
  eprint    = {1908.09791},
  archivePrefix = {arXiv},
  primaryClass  = {cs.LG},
  url       = {https://arxiv.org/abs/1908.09791}
}

@inproceedings{hou2020dynabert,
  title     = {DynaBERT: Dynamic BERT with Adaptive Width and Depth},
  author    = {Hou, Lu and Huang, Zhiqi and Shang, Lifeng and Jiang, Xin and Chen, Xiao and Liu, Qun},
  booktitle = {Advances in Neural Information Processing Systems},
  year      = {2020},
  eprint    = {2004.04037},
  archivePrefix = {arXiv},
  primaryClass  = {cs.CL},
  url       = {https://arxiv.org/abs/2004.04037}
}

@article{jacob2018quantization,
  title         = {Quantization and Training of Neural Networks for Efficient Integer-Arithmetic-Only Inference},
  author        = {Jacob, Benoit and Kligys, Skirmantas and Chen, Bo and Zhu, Menglong and Tang, Matthew and Howard, Andrew and Adam, Hartwig and Kalenichenko, Dmitry},
  journal       = {arXiv preprint arXiv:1712.05877},
  year          = {2017},
  eprint        = {1712.05877},
  archivePrefix = {arXiv},
  primaryClass  = {cs.CV},
  url           = {https://arxiv.org/abs/1712.05877}
}

@inproceedings{esser2020lsq,
  title     = {Learned Step Size Quantization},
  author    = {Esser, Steven K. and McKinstry, Jeffrey L. and Bablani, Deepika and Appuswamy, Rathinakumar and Modha, Dharmendra S.},
  booktitle = {International Conference on Learning Representations (ICLR)},
  year      = {2020},
  eprint    = {1902.08153},
  archivePrefix = {arXiv},
  primaryClass  = {cs.LG},
  url       = {https://openreview.net/forum?id=rkgO66VKDS}
}

@inproceedings{nagel2020adaround,
  title     = {Up or Down? Adaptive Rounding for Post-Training Quantization},
  author    = {Nagel, Markus and van Baalen, Mart and Blankevoort, Tijmen and Welling, Max},
  booktitle = {Proceedings of the 37th International Conference on Machine Learning (ICML)},
  year      = {2020},
  url       = {https://proceedings.mlr.press/v119/nagel20a.html}
}

@inproceedings{li2021brecq,
  title     = {{BRECQ}: Pushing the Limit of Post-Training Quantization by Block Reconstruction},
  author    = {Li, Yuhang and Gong, Ruihao and Tan, Xu and Yang, Yang and Hu, Peng and Zhang, Qi and Yu, Fengwei and Wang, Wei and Gu, Shi},
  booktitle = {International Conference on Learning Representations (ICLR)},
  year      = {2021},
  eprint    = {2102.05426},
  archivePrefix = {arXiv},
  primaryClass  = {cs.LG},
  url       = {https://arxiv.org/abs/2102.05426}
}

@article{yuan2023ptqreliability,
  title         = {Benchmarking the Reliability of Post-training Quantization},
  author        = {Yuan, Zhihang and Liu, Jiawei and Wu, Jiaxiang and Yang, Dawei and Wu, Qiang and Sun, Guangyu and Liu, Wenyu and Wang, Xinggang and Wu, Bingzhe},
  journal       = {arXiv preprint arXiv:2303.13003},
  year          = {2023},
  eprint        = {2303.13003},
  archivePrefix = {arXiv},
  primaryClass  = {cs.LG},
  url           = {https://arxiv.org/abs/2303.13003}
}

@inproceedings{xiao2023smoothquant,
  title     = {SmoothQuant: Accurate and Efficient Post-Training Quantization for Large Language Models},
  author    = {Xiao, Guangxuan and Lin, Ji and Seznec, Mickael and Wu, Hao and Demouth, Julien and Han, Song},
  booktitle = {Proceedings of the 40th International Conference on Machine Learning (ICML)},
  year      = {2023},
  url       = {https://proceedings.mlr.press/v202/xiao23c.html}
}

@inproceedings{dettmers2022llmint8,
  title     = {{LLM.int8()}: 8-bit Matrix Multiplication for Transformers at Scale},
  author    = {Dettmers, Tim and Lewis, Mike and Belkada, Younes and Zettlemoyer, Luke},
  booktitle = {Advances in Neural Information Processing Systems},
  year      = {2022},
  eprint    = {2208.07339},
  archivePrefix = {arXiv},
  primaryClass  = {cs.LG},
  url       = {https://arxiv.org/abs/2208.07339}
}

@article{han2016deepcompression,
  title         = {Deep Compression: Compressing Deep Neural Networks with Pruning, Trained Quantization and Huffman Coding},
  author        = {Han, Song and Mao, Huizi and Dally, William J.},
  journal       = {arXiv preprint arXiv:1510.00149},
  year          = {2015},
  eprint        = {1510.00149},
  archivePrefix = {arXiv},
  primaryClass  = {cs.CV},
  url           = {https://arxiv.org/abs/1510.00149}
}

@article{gale2019stateofsparsity,
  title         = {The State of Sparsity in Deep Neural Networks},
  author        = {Gale, Trevor and Elsen, Erich and Hooker, Sara},
  journal       = {arXiv preprint arXiv:1902.09574},
  year          = {2019},
  eprint        = {1902.09574},
  archivePrefix = {arXiv},
  primaryClass  = {cs.LG},
  url           = {https://arxiv.org/abs/1902.09574}
}

@article{sanh2020movement,
  title         = {Movement Pruning: Adaptive Sparsity by Fine-Tuning},
  author        = {Sanh, Victor and Wolf, Thomas and Rush, Alexander M.},
  journal       = {arXiv preprint arXiv:2005.07683},
  year          = {2020},
  eprint        = {2005.07683},
  archivePrefix = {arXiv},
  primaryClass  = {cs.LG},
  url           = {https://arxiv.org/abs/2005.07683}
}

@inproceedings{frantar2022spdy,
  title     = {{SPDY}: Accurate Pruning with Speedup Guarantees},
  author    = {Frantar, Elias and Alistarh, Dan},
  booktitle = {Proceedings of the 39th International Conference on Machine Learning (ICML)},
  year      = {2022},
  eprint    = {2201.13096},
  archivePrefix = {arXiv},
  primaryClass  = {cs.LG},
  url       = {https://arxiv.org/abs/2201.13096}
}

@article{jaderberg2014lowrank,
  title         = {Speeding up Convolutional Neural Networks with Low Rank Expansions},
  author        = {Jaderberg, Max and Vedaldi, Andrea and Zisserman, Andrew},
  journal       = {arXiv preprint arXiv:1405.3866},
  year          = {2014},
  eprint        = {1405.3866},
  archivePrefix = {arXiv},
  primaryClass  = {cs.CV},
  url           = {https://arxiv.org/abs/1405.3866}
}

@inproceedings{denton2014exploiting,
  title     = {Exploiting Linear Structure Within Convolutional Networks for Efficient Evaluation},
  author    = {Denton, R{\'e}mi and Zaremba, Wojciech and Bruna, Joan and LeCun, Yann and Fergus, Rob},
  booktitle = {Advances in Neural Information Processing Systems},
  year      = {2014},
  eprint    = {1404.0736},
  archivePrefix = {arXiv},
  primaryClass  = {cs.CV},
  url       = {https://arxiv.org/abs/1404.0736}
}

@inproceedings{lebedev2015cp,
  title     = {Speeding-up Convolutional Neural Networks Using Fine-tuned {CP}-Decomposition},
  author    = {Lebedev, Vadim and Ganin, Yaroslav and Rakhuba, Maksim and Oseledets, Ivan and Lempitsky, Victor},
  booktitle = {International Conference on Learning Representations (ICLR)},
  year      = {2015},
  eprint    = {1412.6553},
  archivePrefix = {arXiv},
  primaryClass  = {cs.CV},
  url       = {https://arxiv.org/abs/1412.6553}
}

@inproceedings{lan2020albert,
  title     = {{ALBERT}: A Lite {BERT} for Self-supervised Learning of Language Representations},
  author    = {Lan, Zhenzhong and Chen, Mingda and Goodman, Sebastian and Gimpel, Kevin and Sharma, Piyush and Soricut, Radu},
  booktitle = {International Conference on Learning Representations (ICLR)},
  year      = {2020},
  eprint    = {1909.11942},
  archivePrefix = {arXiv},
  primaryClass  = {cs.CL},
  url       = {https://arxiv.org/abs/1909.11942}
}

@inproceedings{wang2018skipnet,
  title     = {SkipNet: Learning Dynamic Routing in Convolutional Networks},
  author    = {Wang, Xin and Yu, Fisher and Dou, Zi-Yi and Darrell, Trevor and Gonzalez, Joseph E.},
  booktitle = {Proceedings of the European Conference on Computer Vision (ECCV)},
  year      = {2018},
  url       = {https://openaccess.thecvf.com/content_ECCV_2018/html/Xin_Wang_SkipNet_Learning_Dynamic_ECCV_2018_paper.html}
}

@article{graves2016act,
  title         = {Adaptive Computation Time for Recurrent Neural Networks},
  author        = {Graves, Alex},
  journal       = {arXiv preprint arXiv:1603.08983},
  year          = {2016},
  eprint        = {1603.08983},
  archivePrefix = {arXiv},
  primaryClass  = {cs.NE},
  url           = {https://arxiv.org/abs/1603.08983}
}

@inproceedings{weng2018fast,
  title     = {Towards Fast Computation of Certified Robustness for {ReLU} Networks},
  author    = {Weng, Tsui-Wei and Zhang, Huan and Chen, Hongge and Song, Zhao and Hsieh, Cho-Jui and Boning, Duane and Dhillon, Inderjit S. and Daniel, Luca},
  booktitle = {Proceedings of the 35th International Conference on Machine Learning (ICML)},
  year      = {2018},
  eprint    = {1804.09699},
  archivePrefix = {arXiv},
  primaryClass  = {cs.LG},
  url       = {https://arxiv.org/abs/1804.09699}
}

@inproceedings{cohen2019smoothing,
  title     = {Certified Adversarial Robustness via Randomized Smoothing},
  author    = {Cohen, Jeremy and Rosenfeld, Elan and Kolter, Zico},
  booktitle = {Proceedings of the 36th International Conference on Machine Learning (ICML)},
  year      = {2019},
  url       = {https://proceedings.mlr.press/v97/cohen19c.html}
}

@article{hinton2015distilling,
  title         = {Distilling the Knowledge in a Neural Network},
  author        = {Hinton, Geoffrey and Vinyals, Oriol and Dean, Jeff},
  journal       = {arXiv preprint arXiv:1503.02531},
  year          = {2015},
  eprint        = {1503.02531},
  archivePrefix = {arXiv},
  primaryClass  = {cs.LG},
  url           = {https://arxiv.org/abs/1503.02531}
}

@inproceedings{dong2020hawqv2,
  title={HAWQ-V2: Hessian Aware trace-Weighted Quantization of Neural Networks},
  author={Dong, Zhen and Yao, Zhewei and Arfeen, Daiyaan and Gholami, Amir and Mahoney, Michael W. and Keutzer, Kurt},
  booktitle={Advances in Neural Information Processing Systems (NeurIPS)},
  year={2020},
  url={https://proceedings.neurips.cc/paper/2020/hash/d77c703536718b95308130ff2e5cf9ee-Abstract.html}
}

@article{wei2022qdrop,
  title={QDrop: Randomly Dropping Quantization for Extremely Low-bit Post-Training Quantization},
  author={Wei, Xiuying and Gong, Ruihao and Liu, Yingchen and Yu, Fengwei and Wang, Wei and Huang, Junjie and Zhang, Qi and Gu, Shi},
  journal={arXiv preprint arXiv:2203.05740},
  year={2022},
  url={https://arxiv.org/abs/2203.05740}
}

@inproceedings{xu2020gdfq,
  title={Generative Low-bitwidth Data Free Quantization},
  author={Xu, Shoukai and Li, Haokun and Zhuang, Bohan and Liu, Jing and Cao, Jiezhang and Liang, Chuangrun and Tan, Mingkui},
  booktitle={European Conference on Computer Vision (ECCV)},
  year={2020},
  url={https://www.ecva.net/papers/eccv_2020/papers_ECCV/papers/123570001.pdf}
}

@inproceedings{li2020dfqf,
  title={DFQF: Data Free Quantization-aware Fine-tuning},
  author={Li, Bowen and Huang, Kai and Chen, Siang and Xiong, Dongliang and Jiang, Haitian and Claesen, Luc},
  booktitle={Proceedings of The 12th Asian Conference on Machine Learning (ACML)},
  year={2020},
  url={https://proceedings.mlr.press/v129/li20a.html}
}

@inproceedings{zhou2021nm,
  title={Learning N:M Fine-grained Structured Sparse Neural Networks From Scratch},
  author={Zhou, Aojun and Ma, Yukun and Zhu, Junnan and Liu, Jianbo and Zhang, Zhen and Zhang, Li and Zhu, Wenwu and Huang, Gao},
  booktitle={International Conference on Learning Representations (ICLR)},
  year={2021},
  url={https://openreview.net/forum?id=K9bw7vqp_s}
}

@inproceedings{liu2017networksSlimming,
  title={Learning Efficient Convolutional Networks through Network Slimming},
  author={Liu, Zhuang and Li, Jianguo and Shen, Zhiqiang and Huang, Gao and Yan, Shoumeng and Zhang, Changshui},
  booktitle={Proceedings of the IEEE International Conference on Computer Vision (ICCV)},
  year={2017},
  url={https://openaccess.thecvf.com/content_ICCV_2017/papers/Liu_Learning_Efficient_Convolutional_ICCV_2017_paper.pdf}
}

@inproceedings{he2018amc,
  title={AMC: AutoML for Model Compression and Acceleration on Mobile Devices},
  author={He, Yihui and Lin, Ji and Liu, Zhijian and Wang, Hanrui and Li, Li-Jia and Han, Song},
  booktitle={European Conference on Computer Vision (ECCV)},
  year={2018},
  url={https://openaccess.thecvf.com/content_ECCV_2018/papers/Yihui_He_AMC_AutoML_for_ECCV_2018_paper.pdf}
}

@inproceedings{teerapittayanon2016branchynet,
  title={BranchyNet: Fast Inference via Early Exiting from Deep Neural Networks},
  author={Teerapittayanon, Surasak and McDanel, Bradley and Kung, H. T.},
  booktitle={2016 23rd International Conference on Pattern Recognition (ICPR)},
  year={2016},
  url={https://arxiv.org/abs/1709.01686}
}

@article{lebedev2014cp,
  title={Speeding-up Convolutional Neural Networks Using Fine-tuned CP-Decomposition},
  author={Lebedev, Vadim and Ganin, Yaroslav and Rakhuba, Maksim and Oseledets, Ivan and Lempitsky, Victor},
  journal={arXiv preprint arXiv:1412.6553},
  year={2014},
  url={https://arxiv.org/abs/1412.6553}
}

@article{kim2015compressionMobile,
  title={Compression of Deep Convolutional Neural Networks for Fast and Low Power Mobile Applications},
  author={Kim, Yong-Deok and Park, Eunhyeok and Yoo, Sungjoo and Choi, Taelim and Yang, Lu and Shin, Dongjun},
  journal={arXiv preprint arXiv:1511.06530},
  year={2015},
  url={https://arxiv.org/abs/1511.06530}
}

@inproceedings{hu2022lora,
  title={LoRA: Low-Rank Adaptation of Large Language Models},
  author={Hu, Edward J. and Shen, Yelong and Wallis, Phillip and Allen-Zhu, Zeyuan and Li, Yuanzhi and Wang, Shean and Wang, Lu and Chen, Weizhu},
  booktitle={International Conference on Learning Representations (ICLR)},
  year={2022},
  url={https://openreview.net/forum?id=nZeVKeeFYf9}
}

@article{zhang2023adalora,
  title={AdaLoRA: Adaptive Budget Allocation for Parameter-Efficient Fine-Tuning},
  author={Zhang, Qingru and Chen, Minshuo and Bukharin, Alexander and Karampatziakis, Nikos and He, Pengcheng and Cheng, Yu and Chen, Weizhu and Zhao, Tuo},
  journal={arXiv preprint arXiv:2303.10512},
  year={2023},
  url={https://arxiv.org/abs/2303.10512}
}

@inproceedings{valipour2023dylora,
  title={DyLoRA: Parameter Efficient Tuning of Pre-trained Models using Dynamic Search-Free Low-Rank Adaptation},
  author={Valipour, Mojtaba and Rezagholizadeh, Mehdi and Kobyzev, Ivan and Ghodsi, Ali},
  booktitle={Proceedings of the 17th Conference of the European Chapter of the Association for Computational Linguistics (EACL)},
  year={2023},
  url={https://aclanthology.org/2023.eacl-main.239/}
}

@inproceedings{weng2020weightcert,
  title={Towards Certificated Model Robustness Against Weight Perturbations},
  author={Weng, Tsui-Wei and Zhao, Pu and Liu, Sijia and Chen, Pin-Yu and Lin, Xue and Daniel, Luca},
  booktitle={Proceedings of the AAAI Conference on Artificial Intelligence (AAAI)},
  year={2020},
  url={https://ojs.aaai.org/index.php/AAAI/article/view/6105/5961}
}

@article{gowal2018ibp,
  title={On the Effectiveness of Interval Bound Propagation for Training Verifiably Robust Models},
  author={Gowal, Sven and Dvijotham, Krishnamurthy and Stanforth, Robert and Bunel, Rudy and Qin, Chongli and Uesato, Jonathan and Arandjelovic, Relja and Mann, Timothy and Kohli, Pushmeet},
  journal={arXiv preprint arXiv:1810.12715},
  year={2018},
  url={https://arxiv.org/abs/1810.12715}
}

@inproceedings{lin2021integer,
  title={Integer-arithmetic-only Certified Robustness for Quantized Neural Networks},
  author={Lin, Haowen and Lou, Jian and Xiong, Li and Shahabi, Cyrus},
  booktitle={Proceedings of the IEEE/CVF International Conference on Computer Vision (ICCV)},
  year={2021},
  url={https://openaccess.thecvf.com/content/ICCV2021/papers/Lin_Integer-Arithmetic-Only_Certified_Robustness_for_Quantized_Neural_Networks_ICCV_2021_paper.pdf}
}

@inproceedings{deng2009imagenet,
  title     = {ImageNet: A Large-Scale Hierarchical Image Database},
  author    = {Deng, Jia and Dong, Wei and Socher, Richard and Li, Li-Jia and Li, Kai and Fei-Fei, Li},
  booktitle = {Proceedings of the IEEE Conference on Computer Vision and Pattern Recognition (CVPR)},
  year      = {2009},
  url       = {https://www.image-net.org/static_files/papers/imagenet_cvpr09.pdf}
}

@techreport{krizhevsky2009learning,
  title       = {Learning Multiple Layers of Features from Tiny Images},
  author      = {Krizhevsky, Alex},
  institution = {University of Toronto},
  year        = {2009},
  url         = {https://www.cs.toronto.edu/~kriz/learning-features-2009-TR.pdf}
}

@inproceedings{wang2018glue,
  title     = {{GLUE}: A Multi-Task Benchmark and Analysis Platform for Natural Language Understanding},
  author    = {Wang, Alex and Singh, Amanpreet and Michael, Julian and Hill, Felix and Levy, Omer and Bowman, Samuel R.},
  booktitle = {Proceedings of the 2018 EMNLP Workshop {B}lackbox{NLP}},
  year      = {2018},
  url       = {https://aclanthology.org/W18-5446.pdf}
}

@inproceedings{williams2018multinli,
  title     = {A Broad-Coverage Challenge Corpus for Sentence Understanding through Inference},
  author    = {Williams, Adina and Nangia, Nikita and Bowman, Samuel R.},
  booktitle = {Proceedings of the 2018 Conference of the North American Chapter of the Association for Computational Linguistics: Human Language Technologies (NAACL-HLT)},
  year      = {2018},
  url       = {https://aclanthology.org/N18-1101/}
}

@inproceedings{socher2013sentiment,
  title     = {Recursive Deep Models for Semantic Compositionality Over a Sentiment Treebank},
  author    = {Socher, Richard and Perelygin, Alex and Wu, Jean and Chuang, Jason and Manning, Christopher D. and Ng, Andrew and Potts, Christopher},
  booktitle = {Proceedings of the 2013 Conference on Empirical Methods in Natural Language Processing (EMNLP)},
  year      = {2013},
  url       = {https://aclanthology.org/D13-1170/}
}

@inproceedings{merity2017pointer,
  title     = {Pointer Sentinel Mixture Models},
  author    = {Merity, Stephen and Xiong, Caiming and Bradbury, James and Socher, Richard},
  booktitle = {International Conference on Learning Representations (ICLR)},
  year      = {2017},
  url       = {https://arxiv.org/abs/1609.07843}
}

@inproceedings{he2016resnet,
  title     = {Deep Residual Learning for Image Recognition},
  author    = {He, Kaiming and Zhang, Xiangyu and Ren, Shaoqing and Sun, Jian},
  booktitle = {Proceedings of the IEEE Conference on Computer Vision and Pattern Recognition (CVPR)},
  year      = {2016},
  url       = {https://openaccess.thecvf.com/content_cvpr_2016/papers/He_Deep_Residual_Learning_CVPR_2016_paper.pdf}
}

@inproceedings{liu2022convnet,
  title     = {A ConvNet for the 2020s},
  author    = {Liu, Zhuang and Mao, Hanzi and Wu, Chao-Yuan and Feichtenhofer, Christoph and Darrell, Trevor and Xie, Saining},
  booktitle = {Proceedings of the IEEE/CVF Conference on Computer Vision and Pattern Recognition (CVPR)},
  year      = {2022},
  url       = {https://openaccess.thecvf.com/content/CVPR2022/papers/Liu_A_ConvNet_for_the_2020s_CVPR_2022_paper.pdf}
}

@inproceedings{dosovitskiy2021vit,
  title     = {An Image is Worth 16x16 Words: Transformers for Image Recognition at Scale},
  author    = {Dosovitskiy, Alexey and Beyer, Lucas and Kolesnikov, Alexander and Weissenborn, Dirk and Zhai, Xiaohua and Unterthiner, Thomas and Dehghani, Mostafa and Minderer, Matthias and Heigold, Georg and Gelly, Sylvain and Uszkoreit, Jakob and Houlsby, Neil},
  booktitle = {International Conference on Learning Representations (ICLR)},
  year      = {2021},
  url       = {https://arxiv.org/abs/2010.11929}
}

@inproceedings{liu2021swin,
  title     = {Swin Transformer: Hierarchical Vision Transformer using Shifted Windows},
  author    = {Liu, Ze and Lin, Yutong and Cao, Yue and Hu, Han and Wei, Yixuan and Zhang, Zheng and Lin, Stephen and Guo, Baining},
  booktitle = {Proceedings of the IEEE/CVF International Conference on Computer Vision (ICCV)},
  year      = {2021},
  url       = {https://openaccess.thecvf.com/content/ICCV2021/papers/Liu_Swin_Transformer_Hierarchical_Vision_Transformer_Using_Shifted_Windows_ICCV_2021_paper.pdf}
}

@inproceedings{devlin2019bert,
  title     = {{BERT}: Pre-training of Deep Bidirectional Transformers for Language Understanding},
  author    = {Devlin, Jacob and Chang, Ming-Wei and Lee, Kenton and Toutanova, Kristina},
  booktitle = {Proceedings of the 2019 Conference of the North American Chapter of the Association for Computational Linguistics: Human Language Technologies (NAACL-HLT)},
  year      = {2019},
  url       = {https://aclanthology.org/N19-1423.pdf}
}

@article{liu2019roberta,
  title         = {RoBERTa: A Robustly Optimized {BERT} Pretraining Approach},
  author        = {Liu, Yinhan and Ott, Myle and Goyal, Naman and Du, Jingfei and Joshi, Mandar and Chen, Danqi and Levy, Omer and Lewis, Mike and Zettlemoyer, Luke and Stoyanov, Veselin},
  journal       = {arXiv preprint arXiv:1907.11692},
  year          = {2019},
  url           = {https://arxiv.org/abs/1907.11692}
}

@article{sanh2019distilbert,
  title         = {DistilBERT, a distilled version of BERT: smaller, faster, cheaper and lighter},
  author        = {Sanh, Victor and Debut, Lysandre and Chaumond, Julien and Wolf, Thomas},
  journal       = {arXiv preprint arXiv:1910.01108},
  year          = {2019},
  url           = {https://arxiv.org/abs/1910.01108}
}

@article{zhang2024tinylamma,
  title         = {TinyLlama: An Open-Source Small Language Model},
  author        = {Zhang, Peiyuan and Zeng, Guangtao and Wang, Tianduo and Lu, Wei},
  journal       = {arXiv preprint arXiv:2401.02385},
  year          = {2024},
  url           = {https://arxiv.org/abs/2401.02385}
}

@inproceedings{wang2019haq,
  title     = {HAQ: Hardware-Aware Automated Quantization With Mixed Precision},
  author    = {Wang, Kuan and Liu, Zhijian and Lin, Yujun and Lin, Ji and Han, Song},
  booktitle = {Proceedings of the IEEE/CVF Conference on Computer Vision and Pattern Recognition (CVPR)},
  year      = {2019},
  url       = {https://openaccess.thecvf.com/content_CVPR_2019/papers/Wang_HAQ_Hardware-Aware_Automated_Quantization_With_Mixed_Precision_CVPR_2019_paper.pdf}
}

@inproceedings{han2015learningconnections,
  title     = {Learning both Weights and Connections for Efficient Neural Network},
  author    = {Han, Song and Pool, Jeff and Tran, John and Dally, William},
  booktitle = {Advances in Neural Information Processing Systems (NeurIPS)},
  year      = {2015},
  url       = {https://arxiv.org/abs/1506.02626}
}

@article{frantar2023sparsegpt,
  title         = {SparseGPT: Massive Language Models Can Be Accurately Pruned in One-Shot},
  author        = {Frantar, Elias and Alistarh, Dan},
  journal       = {arXiv preprint arXiv:2301.00774},
  year          = {2023},
  url           = {https://arxiv.org/abs/2301.00774}
}

@article{shin2019kdquant,
  title         = {Knowledge distillation for optimization of quantized deep neural networks},
  author        = {Shin, Sungho and Boo, Yoonho and Sung, Wonyong},
  journal       = {arXiv preprint arXiv:1909.01688},
  year          = {2019},
  url           = {https://arxiv.org/abs/1909.01688}
}

@inproceedings{choi2020dfqakd,
  title     = {Data-Free Network Quantization With Adversarial Knowledge Distillation},
  author    = {Choi, Yoojin and Yu, Jihwan and Wook Shin, Young and Lee, Jongseok and Kim, Minsu},
  booktitle = {Proceedings of the IEEE/CVF Conference on Computer Vision and Pattern Recognition Workshops (CVPRW)},
  year      = {2020},
  url       = {https://openaccess.thecvf.com/content_CVPRW_2020/papers/w40/Choi_Data-Free_Network_Quantization_With_Adversarial_Knowledge_Distillation_CVPRW_2020_paper.pdf}
}

@inproceedings{xiao2018gandiva,
  author    = {Wencong Xiao and Romil Bhardwaj and Ramachandran Ramjee and Muthian Sivathanu and Nipun Kwatra and Zhenhua Han and Pratyush Patel and Xuan Peng and Hanyu Zhao and Quanlu Zhang and Fan Yang and Lidong Zhou},
  title     = {Gandiva: Introspective Cluster Scheduling for Deep Learning},
  booktitle = {13th USENIX Symposium on Operating Systems Design and Implementation (OSDI 18)},
  year      = {2018},
  pages     = {595--610},
  publisher = {USENIX Association}
}

@inproceedings{gu2019tiresias,
  author    = {Juncheng Gu and Mosharaf Chowdhury and Kang G. Shin and Yibo Zhu and Myeongjae Jeon and Junjie Qian and Hongqiang Liu and Chuanxiong Guo},
  title     = {Tiresias: A GPU Cluster Manager for Distributed Deep Learning},
  booktitle = {16th USENIX Symposium on Networked Systems Design and Implementation (NSDI 19)},
  year      = {2019},
  publisher = {USENIX Association}
}

@article{benoittcattin2020thermal,
  title={Impact of thermal throttling on long-term visual inference in a CPU-based edge device},
  author={Benoit-Cattin, Th{\'e}o and Velasco-Montero, Delia and Fern{\'a}ndez-Berni, Jorge},
  journal={Electronics},
  volume={9},
  number={12},
  pages={2106},
  year={2020},
  publisher={MDPI}
}

@inproceedings{lin2023geardvfs,
  title={A workload-aware DVFS robust to concurrent tasks for mobile devices},
  author={Lin, Chengdong and Wang, Kun and Li, Zhenjiang and Pu, Yu},
  booktitle={Proceedings of the 29th Annual International Conference on Mobile Computing and Networking},
  pages={1--16},
  year={2023}
}

@inproceedings{boroumand2021mensa,
  title={Google neural network models for edge devices: Analyzing and mitigating machine learning inference bottlenecks},
  author={Boroumand, Amirali and Ghose, Saugata and Akin, Berkin and Narayanaswami, Ravi and Oliveira, Geraldo F and Ma, Xiaoyu and Shiu, Eric and Mutlu, Onur},
  booktitle={2021 30th International Conference on Parallel Architectures and Compilation Techniques (PACT)},
  pages={159--172},
  year={2021},
  organization={IEEE}
}

@article{tolstikhin2021mlpmixer,
  title={Mlp-mixer: An all-mlp architecture for vision},
  author={Tolstikhin, Ilya O and Houlsby, Neil and Kolesnikov, Alexander and Beyer, Lucas and Zhai, Xiaohua and Unterthiner, Thomas and Yung, Jessica and Steiner, Andreas and Keysers, Daniel and Uszkoreit, Jakob and others},
  journal={Advances in neural information processing systems},
  volume={34},
  pages={24261--24272},
  year={2021}
}

@article{raffel2020t5,
  title={Exploring the limits of transfer learning with a unified text-to-text transformer},
  author={Raffel, Colin and Shazeer, Noam and Roberts, Adam and Lee, Katherine and Narang, Sharan and Matena, Michael and Zhou, Yanqi and Li, Wei and Liu, Peter J},
  journal={Journal of machine learning research},
  volume={21},
  number={140},
  pages={1--67},
  year={2020}
}

@article{micikevicius2022fp8,
  title={Fp8 formats for deep learning},
  author={Micikevicius, Paulius and Stosic, Dusan and Burgess, Neil and Cornea, Marius and Dubey, Pradeep and Grisenthwaite, Richard and Ha, Sangwon and Heinecke, Alexander and Judd, Patrick and Kamalu, John and others},
  journal={arXiv preprint arXiv:2209.05433},
  year={2022}
}

@article{parashar2017scnn,
  title={SCNN: An accelerator for compressed-sparse convolutional neural networks},
  author={Parashar, Angshuman and Rhu, Minsoo and Mukkara, Anurag and Puglielli, Antonio and Venkatesan, Rangharajan and Khailany, Brucek and Emer, Joel and Keckler, Stephen W and Dally, William J},
  journal={ACM SIGARCH computer architecture news},
  volume={45},
  number={2},
  pages={27--40},
  year={2017},
  publisher={ACM New York, NY, USA}
}

@misc{onnx,
  author = {{ONNX Community}},
  title  = {{ONNX}: Open Neural Network Exchange},
  year   = {2017},
  howpublished = {\url{https://onnx.ai/}}
}

@misc{tensorrt,
  author = {{NVIDIA}},
  title  = {{TensorRT}: NVIDIA Inference Optimizer and Runtime},
  year   = {2017},
  howpublished = {\url{https://developer.nvidia.com/tensorrt}}
}

@inproceedings{reddi2020mlperfinference,
  title={Mlperf inference benchmark},
  author={Reddi, Vijay Janapa and Cheng, Christine and Kanter, David and Mattson, Peter and Schmuelling, Guenther and Wu, Carole-Jean and Anderson, Brian and Breughe, Maximilien and Charlebois, Mark and Chou, William and others},
  booktitle={2020 ACM/IEEE 47th Annual International Symposium on Computer Architecture (ISCA)},
  pages={446--459},
  year={2020},
  organization={IEEE}
}

@article{choi2018pact,
  title={Pact: Parameterized clipping activation for quantized neural networks},
  author={Choi, Jungwook and Wang, Zhuo and Venkataramani, Swagath and Chuang, Pierce I-Jen and Srinivasan, Vijayalakshmi and Gopalakrishnan, Kailash},
  journal={arXiv preprint arXiv:1805.06085},
  year={2018}
}

@inproceedings{nagel2019dfq,
  title={Data-free quantization through weight equalization and bias correction},
  author={Nagel, Markus and Baalen, Mart van and Blankevoort, Tijmen and Welling, Max},
  booktitle={Proceedings of the IEEE/CVF international conference on computer vision},
  pages={1325--1334},
  year={2019}
}

@misc{zhou2016dorefa,
  author       = {Shuchang Zhou and others},
  title        = {DoReFa-Net: Training Low Bitwidth Convolutional Neural Networks with Low Bitwidth Gradients},
  year         = {2016},
  eprint       = {1606.06160},
  archivePrefix= {arXiv},
  primaryClass = {cs.CV}
}

@inproceedings{dong2019hawq,
  title={Hawq: Hessian aware quantization of neural networks with mixed-precision},
  author={Dong, Zhen and Yao, Zhewei and Gholami, Amir and Mahoney, Michael W and Keutzer, Kurt},
  booktitle={Proceedings of the IEEE/CVF international conference on computer vision},
  pages={293--302},
  year={2019}
}

@article{lou2020autoq,
  title={Autoq: Automated kernel-wise neural network quantization},
  author={Lou, Qian and Guo, Feng and Liu, Lantao and Kim, Minje and Jiang, Lei},
  journal={arXiv preprint arXiv:1902.05690},
  year={2019}
}

@article{frantar2022gptq,
  title={Gptq: Accurate post-training quantization for generative pre-trained transformers},
  author={Frantar, Elias and Ashkboos, Saleh and Hoefler, Torsten and Alistarh, Dan},
  journal={arXiv preprint arXiv:2210.17323},
  year={2022}
}

@inproceedings{luo2017thinet,
  author    = {Jian{-}Hao Luo and Jianxin Wu and Weiyao Lin},
  title     = {{ThiNet}: A Filter Level Pruning Method for Deep Neural Network Compression},
  booktitle = {Proceedings of the IEEE International Conference on Computer Vision (ICCV)},
  year      = {2017}
}

@article{lee2019snip,
  title={Snip: Single-shot network pruning based on connection sensitivity},
  author={Lee, Namhoon and Ajanthan, Thalaiyasingam and Torr, Philip HS},
  journal={arXiv preprint arXiv:1810.02340},
  year={2018}
}

@article{frankle2019lth,
  title={The lottery ticket hypothesis: Finding sparse, trainable neural networks},
  author={Frankle, Jonathan and Carbin, Michael},
  journal={arXiv preprint arXiv:1803.03635},
  year={2018}
}

@article{louizos2018l0,
  title={Learning sparse neural networks through $ L\_0 $ regularization},
  author={Louizos, Christos and Welling, Max and Kingma, Diederik P},
  journal={arXiv preprint arXiv:1712.01312},
  year={2017}
}

@inproceedings{evci2020rigl,
  title={Rigging the lottery: Making all tickets winners},
  author={Evci, Utku and Gale, Trevor and Menick, Jacob and Castro, Pablo Samuel and Elsen, Erich},
  booktitle={International conference on machine learning},
  pages={2943--2952},
  year={2020},
  organization={PMLR}
}

@article{novikov2015tensorizing,
  title={Tensorizing neural networks},
  author={Novikov, Alexander and Podoprikhin, Dmitrii and Osokin, Anton and Vetrov, Dmitry P},
  journal={Advances in neural information processing systems},
  volume={28},
  year={2015}
}

@article{wang2020linformer,
  title={Linformer: Self-attention with linear complexity},
  author={Wang, Sinong and Li, Belinda Z and Khabsa, Madian and Fang, Han and Ma, Hao},
  journal={arXiv preprint arXiv:2006.04768},
  year={2020}
}

@article{dettmers2023qlora,
  title={Qlora: Efficient finetuning of quantized llms},
  author={Dettmers, Tim and Pagnoni, Artidoro and Holtzman, Ari and Zettlemoyer, Luke},
  journal={Advances in neural information processing systems},
  volume={36},
  pages={10088--10115},
  year={2023}
}

@article{huang2018msdnet,
  title={Multi-scale dense networks for resource efficient image classification},
  author={Huang, Gao and Chen, Danlu and Li, Tianhong and Wu, Felix and Van Der Maaten, Laurens and Weinberger, Kilian Q},
  journal={arXiv preprint arXiv:1703.09844},
  year={2017}
}

@article{rao2021dynamicvit,
  title={Dynamicvit: Efficient vision transformers with dynamic token sparsification},
  author={Rao, Yongming and Zhao, Wenliang and Liu, Benlin and Lu, Jiwen and Zhou, Jie and Hsieh, Cho-Jui},
  journal={Advances in neural information processing systems},
  volume={34},
  pages={13937--13949},
  year={2021}
}

@article{fedus2021switch,
  title={Switch transformers: Scaling to trillion parameter models with simple and efficient sparsity},
  author={Fedus, William and Zoph, Barret and Shazeer, Noam},
  journal={Journal of Machine Learning Research},
  volume={23},
  number={120},
  pages={1--39},
  year={2022}
}

@article{zhang2018crown,
  title={Efficient neural network robustness certification with general activation functions},
  author={Zhang, Huan and Weng, Tsui-Wei and Chen, Pin-Yu and Hsieh, Cho-Jui and Daniel, Luca},
  journal={Advances in neural information processing systems},
  volume={31},
  year={2018}
}
\bibliographystyle{icml2026}

\newpage
\appendix
\onecolumn
\section{Extended Background \& Related Work}
\label{AppA}

\subsection*{A.1 \quad Expanded motivation \& ecosystem needs} Modern ML deployment spans a jagged landscape of hardware, latency targets, and reliability constraints that change from hour to hour, user to user, and app to app; the same model may run in a multiplexed cloud service one minute, on a thermally throttled edge accelerator the next, and finally on a battery-constrained mobile CPU where memory bandwidth dominates FLOPs.\citep{xiao2018gandiva,gu2019tiresias,benoittcattin2020thermal,lin2023geardvfs,boroumand2021mensa} In such settings, the dominant pain points arise from \emph{rigidity of exported artifacts}, \emph{budget uncertainty at runtime}, and \emph{lack of dependable guardrails when compressing}. 

First, classical pipelines materialize one compressed checkpoint per operating point (e.g., an 8-bit QAT model for server, a 4-bit PTQ variant for mobile, a pruned variant for an NPU). This multiplies storage, complicates A/B and rollback, fractures monitoring, and forces brittle routing logic that is sensitive to device idiosyncrasies and kernel availability. Second, real-world budgets are stochastic: co-tenancy and scheduler noise on GPUs, temperature and DVFS on edge devices, user-visible frame deadlines in interactive workloads, and on-device memory pressure that shifts with other applications. Operators therefore need \emph{elasticity} at inference time: a knob that moves the model along the accuracy–latency–size frontier without retraining or re-exporting, and without violating hardware constraints (kernel shapes, quantization formats, alignment). 

Third, production reliability increasingly demands \emph{predictable failure modes}—particularly when compression trades accuracy for efficiency. SLAs and safety reviews ask not only “how fast” and “how accurate,” but “how far could the output shift when we trim 20\% compute?” Today’s answers are empirical and often brittle; formal robustness work rarely translates into actionable deployment knobs, while many high-performing compression methods offer limited guarantees. The ecosystem trendlines intensify these pressures: model families are larger and more heterogeneous (CNNs, ViTs, MLP-Mixers, encoder/decoder LMs),\citep{dosovitskiy2021vit,tolstikhin2021mlpmixer,raffel2020t5} accelerators expose diverse datatypes (INT8, INT4, FP8, mixed per-channel schemes)\citep{micikevicius2022fp8} and sparse/low-rank kernels that are highly vendor-specific,\citep{parashar2017scnn} and privacy/latency requirements push more inference to the edge. 

Operators need a \emph{single, portable checkpoint} that can be \emph{steered at test time} to hit budget targets on unfamiliar devices, with \emph{monotone behavior} (a larger budget never hurts accuracy) and \emph{telemetry hooks} that quantify risk. From a tooling viewpoint, the artifact should integrate with common export/compile steps (ONNX/TensorRT, mobile runtimes),\citep{onnx,tensorrt} snap to discrete hardware profiles to avoid kernel gaps, and surface lightweight \emph{certificates} that bound induced drift under admissible compressions. Finally, evaluation itself must evolve: heterogeneous-device Pareto curves with p50/p90 latencies, energy per inference, memory footprint including activation buffers, and \emph{budget-respect rates} (how often a requested profile meets its latency/energy target) are as important as single-number accuracy.\citep{reddi2020mlperfinference} In short, the ecosystem needs compression that is \emph{adaptive, certifiable, hardware-aware, and operationally simple}, converting compression from an offline engineering fork into an online control plane aligned with production SLOs. 

\subsection*{A.2 \quad Detailed survey (quantization, pruning, low-rank, dynamic nets, verification)} 
\textbf{Quantization.} Post-training quantization (PTQ) calibrates scales/zero-points on a held-out set and is attractive for speed and simplicity, but its accuracy can degrade at ultra-low precision or under distribution shift; mitigations include per-channel scales, activation clipping, bias correction, and outlier-aware schemes that carve high-variance channels into higher precision. Quantization-aware training (QAT) replaces hard rounding with STE-based surrogates to learn scale/clipping parameters, generally improving robustness but at the cost of additional training and a fixed exported bit-width. Mixed-precision methods cast bit allocation as a search or differentiable relaxation (reward/gradients from latency proxies) to distribute bits across layers; deployment friction arises when the chosen mix does not map cleanly to vendor kernels, necessitating profile pruning. Recent PTQ/QAT hybrids add knowledge distillation and cross-layer equalization; they push the pareto frontier but still \emph{freeze} precision at export and provide no explicit bound on output drift.\citep{choi2018pact,nagel2019dfq,zhou2016dorefa,dong2019hawq,lou2020autoq,frantar2022gptq} 

\paragraph{Pruning and sparsity.} Unstructured pruning achieves high parameter sparsity and small on-disk size, but real speedups require fine-grained sparse kernels with nontrivial overheads; structured or channel pruning yields reliable latency gains by shrinking whole filters/projections, though it often needs iterative re-training and careful schedules to avoid accuracy cliffs. Movement- and magnitude-based criteria, L0/L1 relaxations, and lottery-ticket style rewinds populate the design space; dynamic sparsity adds input adaptivity but complicates compilation and predictability.\citep{han2016deepcompression,luo2017thinet,lee2019snip,frankle2019lth,louizos2018l0,evci2020rigl} 

\paragraph{Low-rank and tensor factorization.} SVD/Tucker/CP decompositions reduce parameters and memory traffic in dense and convolutional layers; in Transformers, factorizations applied to projection matrices and MLP blocks trade rank for accuracy. LoRA-style adapters inject low-rank updates during fine-tuning, and low-rank \emph{replacement} compresses the base weights; most approaches fix rank offline and re-export per rank, which clashes with runtime adaptivity and kernel availability. Joint rank–precision allocation is comparatively underexplored: quantization interacts with rank because truncation changes dynamic ranges and singular spectra, which in turn affects scale selection and error propagation.\citep{novikov2015tensorizing,wang2020linformer,dettmers2023qlora} 

\paragraph{Dynamic networks.} Early exiting, token/channel dropping, and input-conditional routing tailor compute to difficulty, improving average latency. However, they require architectural hooks, retraining, and careful calibration to avoid pathological exits; guarantees are typically statistical (expected cost) rather than \emph{per-budget} deterministic. Moreover, dynamic policies can adversarially interact with batching, caching, or compiler fusions, yielding unstable tail latencies.\citep{huang2018msdnet,rao2021dynamicvit,fedus2021switch} 

\paragraph{Verification and certificates.} Robustness certificates bound output change under input or parameter perturbations (e.g., Lipschitz, interval/zonotope bounds, randomized smoothing). Parameter-perturbation guarantees—relevant for compression—bound the effect of quantization noise, pruning, or low-rank truncation, but many are conservative or expensive to compute at scale, and few are wired into a deployment-time control loop. Practical adoption is hindered by the gap between math-friendly assumptions and production details like residual connections, normalization, fused kernels, and calibration drift.\citep{zhang2018crown,cohen2019smoothing,weng2018fast} 

\paragraph{Positioning.} Against this backdrop, a practical method should: 1) \emph{train once} yet expose a \emph{continuous} control that snaps to \emph{discrete hardware profiles} at inference; 2) \emph{jointly} reason about rank and precision because their errors compound and their costs couple to memory bandwidth and kernel shapes; 3 provide a \emph{fast, layerwise certificate} that upper-bounds logit drift for the selected profile and can be surfaced to monitoring; and (iv) preserve \emph{monotonicity in budget} so operators can raise a target without risking accuracy regressions. The prevailing literature supplies powerful pieces—accurate PTQ/QAT at fixed bits, effective pruning schedules, strong low-rank adapters, and rigorous but heavy certificates—but rarely assembles them into an elastic, certifiable, and hardware-conscious pipeline that speaks the language of production SLOs. T3C is designed to bridge this gap by sharing parameters across ranks, tying bit-width to rank in a deployable way, and attaching a certificate that is cheap enough for training-time regularization and export-time reporting, thereby turning compression into an operational control rather than an offline fork.

\section{Method Details}
\label{AppB}
\subsection*{B.1 \quad Notation \& tensor decompositions (SVD/Tucker/CP variants)}
\paragraph{Basic notation.}
We consider a feed-forward network with layers indexed by $\ell\in\{1,\dots,L\}$.
For a dense (fully-connected) layer $\ell$, let $W_\ell\in\mathbb{R}^{m_\ell\times n_\ell}$ map $a_{\ell-1}\in\mathbb{R}^{n_\ell}$ to $h_\ell=W_\ell a_{\ell-1}$ (biases omitted for brevity); activations may include normalization and nonlinearity $h_\ell\mapsto a_\ell$.
For a convolutional kernel, $W_\ell\in\mathbb{R}^{C_{\mathrm{out}}\times C_{\mathrm{in}}\times h\times w}$ uses stride/padding as in the base model.
Given input $x$, the full model produces logits $z=f(x)$; under a budget profile $k$ with per-layer rank/bit $(k_\ell,q_\ell)$, we denote the compressed model by $\tilde f_k(x)$ with logits $\tilde z$.

\paragraph{SVD (matrix) factorization.}
For $W\in\mathbb{R}^{m\times n}$, the rank-$r$ truncated SVD is $W_r=U_{:,1:r}\Sigma_{1:r}V_{:,1:r}^{\top}$ with singular values $\sigma_1\ge\dots\ge\sigma_r$.
We maintain factors up to $k_{\max}$ and expose a differentiable top-$k\!\le\!k_{\max}$ selection (Sec.~B.2).
We employ two numerically stable parameterizations:
(i) \emph{spectral form} $U,\Sigma,V$ with orthogonality constraints enforced by implicit re-orthogonalization via Householder updates;
(ii) \emph{product form} $A B^{\top}$, $A\in\mathbb{R}^{m\times k_{\max}}$, $B\in\mathbb{R}^{n\times k_{\max}}$, with a spectral penalty encouraging $A$ and $B$ to approximate singular directions.
Spectral form yields direct control of the residual norm $\|W-W_k\|_2=\sigma_{k+1}$ and $\|W-W_k\|_F=(\sum_{i>k}\sigma_i^2)^{1/2}$, useful for certificates.

\paragraph{Tucker-2 for convolution.}
For $W\in\mathbb{R}^{C_o\times C_i\times h\times w}$ we use channel-only Tucker-2:
\begin{equation}
W \approx (U_o, U_i) \cdot G,\quad
U_o\in\mathbb{R}^{C_o\times r_o},\; U_i\in\mathbb{R}^{C_i\times r_i},\;
G\in\mathbb{R}^{r_o\times r_i\times h\times w}.
\end{equation}
The effective forward is $\mathrm{Conv}(U_o\,\,\mathrm{Conv}(G,\, U_i^\top\,\cdot))$, which compiles to three kernels (pointwise $1{\times}1$, spatial $h{\times}w$, pointwise $1{\times}1$).
Budget $k$ maps to $(r_o(k),r_i(k))$ via a monotone schedule.

\paragraph{CP for depthwise/attention blocks.}
For depthwise-like or MLP projection weights with strong separability, a CP rank-$r$ factorization $W\approx \sum_{j=1}^r a^{(1)}_j\otimes a^{(2)}_j$ reduces memory traffic; in attention, we optionally share a single $k$ across $\{W_Q,W_K,W_V,W_O\}$ to avoid head imbalance.

\paragraph{Identifiability and conditioning.}
Tucker and CP are non-unique up to scale/permutation. We eliminate degeneracies with per-factor $\ell_2$ normalization and a permutation-fixing rule (descending factor norms). To prevent ill-conditioning at small ranks, we regularize the spectrum by $\sum_i \max(0,\sigma_i-\sigma_{i+1}-\delta)$ to keep gaps from collapsing (small $\delta$).

\subsection*{B.2 \quad Loss terms, annealing schedules, $\lambda$ sweeps}
\paragraph{Total objective (expanded).}
For a minibatch $\mathcal{B}$ and sampled budget profile $k$, the loss is
\begin{align}
\mathcal{L} &=
\underbrace{\frac{1}{|\mathcal{B}|}\sum_{(x,y)\in\mathcal{B}} \mathrm{CE}(f_{\text{full}}(x),y)}_{\text{task (full)}}\;+\;
\lambda_{\mathrm{SD}}\,\underbrace{\frac{1}{|\mathcal{B}|}\sum_{x\in\mathcal{B}} \mathrm{KL}\big(p_{\text{full}}(x)\|p_k(x)\big)}_{\text{self-distill}}\nonumber\\
&\quad +\lambda_{\mathrm{AUG}}\,\underbrace{\mathbb{E}_{\tilde x\sim \mathcal{T}(x)}\mathrm{KL}\big(p_{\text{full}}(\tilde x)\|p_k(\tilde x)\big)}_{\text{augmentation consistency}}
+\lambda_{\mathrm{CERT}}\,\underbrace{\max\!\big(0,\,\hat\Delta(k)-\epsilon\big)}_{\text{certificate penalty}}
+\lambda_{\mathrm{BUD}}\,\underbrace{\mathrm{Cost}(k;b)}_{\text{budget proxy}},
\label{eq:full-loss-appendix}
\end{align}
with $p(\cdot)=\mathrm{softmax}(z/T)$ (temperature $T{=}1$ unless specified). The \emph{task (full)} term trains the high-rank view ($k\!=\!k_{\max}$ or a high-rank proxy), anchoring teacher quality. The two KL terms align compressed predictions to the full model on raw and lightly augmented inputs. The certificate penalty uses the bound from Sec.~\ref{sec:certificate} (main paper) with running estimates. The budget proxy is the expected normalized latency/energy/size against a target.

\paragraph{Annealing schedules.}
\emph{Top-$k$ temperature.} For the Gumbel-Top-$k$ mask logits $g_i$, we use temperature $\tau_t=\max(\tau_{\min},\,\tau_0\cdot \gamma^{t/T_{\mathrm{anneal}}})$ with $\gamma\in(0,1)$, annealed over $T_{\mathrm{anneal}}$ steps, then held.
\emph{Rank sampling.} Early curriculum: sample $k$ from a wide Beta$(\alpha{=}0.75,\beta{=}0.75)$ on $[k_{\min},k_{\max}]$, then bias toward deployment-relevant profiles by mixing a categorical over discrete profiles $\{k^{(j)}\}_j$.
\emph{Coefficient ramps.} We use linear warmups for $\lambda_{\mathrm{SD}},\lambda_{\mathrm{AUG}},\lambda_{\mathrm{CERT}}$ over the first 10–20\% of training to avoid early over-regularization.

\paragraph{\boldmath $\lambda$ sweeps and stability.}
We recommend a coarse-to-fine sweep: fix $\lambda_{\mathrm{BUD}}$ to meet a target utilization (latency/energy) on the validation device, then search $\lambda_{\mathrm{CERT}}$ to cap violation rates (fraction of samples where $\hat\Delta(k)>\epsilon$), and finally tune $\lambda_{\mathrm{SD}}$ and $\lambda_{\mathrm{AUG}}$ to recover accuracy at tight budgets. In practice, a good starting box is
\begin{equation}
\lambda_{\mathrm{SD}}\in[0.3,1.0],\quad
\lambda_{\mathrm{AUG}}\in[0.1,0.5],\quad
\lambda_{\mathrm{CERT}}\in[0.05,0.5],\quad
\lambda_{\mathrm{BUD}}\in[0.1,1.0].
\end{equation}
We monitor (i) monotonicity (accuracy should be non-decreasing in budget), (ii) \% budget violations, and (iii) calibration error between $\hat\Delta(k)$ and observed drift.

\subsection*{B.3 \quad Quantizer definitions, calibration, rounding tricks}
\paragraph{Uniform affine quantization.}
For a real tensor $T$, $q$-bit symmetric per-tensor quantization uses scale $s>0$ and integer grid $\mathcal{G}_q=\{-2^{q-1}\!+\!1,\dots,2^{q-1}\!-\!1\}$:
\begin{equation}
\mathrm{Quantize}_q(T) \;=\; s\,\mathrm{clip}\!\left(\left\lfloor \frac{T}{s} \right\rceil,\;\min\mathcal{G}_q,\;\max\mathcal{G}_q\right),\qquad
s=\frac{\max(|T|)}{2^{q-1}-1}.
\end{equation}
We also use per-channel scales $s_c$ for matrices along output channels (rows) and for conv along $C_{\mathrm{out}}$.

\paragraph{Straight-through estimator (STE).}
In backprop, we treat $\partial \lfloor u \rceil/\partial u \approx 1$ inside the clipping range and $0$ outside. For scale $s$, we learn $s$ (log-parameterized) with gradient
$\frac{\partial\, \mathrm{Quantize}_q(T)}{\partial s}\approx
\left(\lfloor T/s \rceil - T/s\right)$ inside range, encouraging scale to match the signal range.

\paragraph{Rounding variants.}
\emph{Stochastic rounding} $\lfloor u \rceil_{\mathrm{stoch}}= \lfloor u \rfloor$ with prob $1-(u-\lfloor u \rfloor)$ else $\lceil u \rceil$, used at training-time to reduce bias at low $q$; disabled at export for determinism.
\emph{Bias correction} adds a folded offset $\delta=\mathbb{E}[T-\mathrm{Quantize}_q(T)]$ (per-channel) to the dequantized tensor during calibration and is then fused into bias parameters.
\emph{Outlier-aware splitting} routes a small fraction $\rho$ of largest-magnitude channels to $q{+}\Delta q$ bits when hardware supports mixed precision within a layer; we expose this as a profile option so the controller can assign higher bits to outlier groups.

\paragraph{Calibration.}
We compute $s$ (and optionally zero-point for asymmetric quantization) from a calibration set $\mathcal{C}$ via either:
(i) \emph{max-range} (as above), or
(ii) \emph{percentile clipping} $s = \mathrm{percentile}_{p}(|T|)/(2^{q-1}-1)$ with $p\in[99.0,99.99]$ tuned to minimize validation KL between float and quantized outputs of the layer.
We maintain EMA statistics of $s$ during training to stabilize deployment scales.

\paragraph{Rank-tied bit allocation.}
We use a monotone map $q(k)=\min\{q_{\max},\, \lfloor a \log k + b \rfloor\}$ per factor group with small per-factor offsets (e.g., $q_U=q(k)\!+\!1$, $q_G=q(k)$, $q_V=q(k)\!+\!1$) ensuring higher rank receives at least as many bits. This interacts favorably with spectral decay: smaller ranks truncate more energy and benefit less from extra precision.

\paragraph{Lemma (expected dequantization error under stochastic rounding).}
Let $X$ be a scalar with $|X|\le M$ and fixed scale $s$. With stochastic rounding on the grid $s\mathcal{G}_q$ and no clipping, the dequantization error $\varepsilon=\mathrm{Quantize}_q(X)-X$ satisfies $\mathbb{E}[\varepsilon\,|\,X]=0$ and $\mathrm{Var}(\varepsilon\,|\,X)\le s^2/4$.
\emph{Proof.} Conditioning on $u=X/s$, the rounding picks $\lfloor u \rfloor$ with prob $1-(u-\lfloor u \rfloor)$ and $\lceil u \rceil$ otherwise; the expectation equals $u$. Multiply by $s$ to get unbiasedness; the conditional variance of a Bernoulli with step size $s$ is bounded by $s^2/4$.\hfill$\square$

\subsection*{B.4 \quad Controller architectures \& training (relaxations, baselines)}
\paragraph{Budget tokenization.}
A budget $b$ encodes target latency/energy/size and optionally a device id. We embed $b$ as
$e_b=\mathrm{Embed}_{\text{dev}}(\mathrm{id}) \oplus \phi(\mathrm{lat},\mathrm{energy},\mathrm{size})$
where $\phi$ is a learned MLP on normalized scalars; $\oplus$ denotes concatenation.

\paragraph{Policy head.}
We use a shared two-layer MLP $\pi_\phi$ that outputs per-layer logits for discrete profiles and, optionally, a continuous proposal projected onto those profiles:
\begin{equation}
\hat{k}_\ell, \hat{q}_\ell = \mathrm{Proj}\!\left(W_2\,\sigma(W_1[e_b \oplus s(x)])\right),\qquad
\mathrm{Proj}:\mathbb{R}^{2}\!\to \mathcal{P}_\ell\subset\{(k,q)\}.
\end{equation}
Here $s(x)$ is an optional input summary (e.g., pooled penultimate activations averaged over the batch). $\mathcal{P}_\ell$ is a layer-specific menu of hardware-aligned $(k,q)$ pairs.

\paragraph{Discrete relaxations.}
We use Gumbel-Softmax over $\mathcal{P}_\ell$ with temperature $\tau$ and the straight-through trick to obtain a one-hot (profile index) in the forward pass and a soft distribution in the backward pass. This preserves end-to-end differentiability while training the controller jointly with factors and quantizers.

\paragraph{Training objectives for the controller.}
The controller is trained implicitly by the total loss (Eq.~\ref{eq:full-loss-appendix}); gradients propagate through the differentiable relaxation and the budget/certificate penalties. When using purely discrete selection (no relaxation), we add a REINFORCE term:
\begin{equation}
\nabla_\phi \mathbb{E}_{\pi_\phi}\big[-\mathcal{L}\big] \approx \mathbb{E}\big[(-\mathcal{L}-b)\,\nabla_\phi\log \pi_\phi\big],
\end{equation}
with a learned baseline $b$ (value head) to reduce variance. In practice, the Gumbel-Softmax + straight-through suffices, switching to pure argmax after the anneal.

\paragraph{Monotonicity enforcement.}
We ensure that higher budgets cannot yield smaller ranks/bits via isotonic constraints: for two budget tokens $b_1\prec b_2$ (componentwise), we add a hinge
\begin{equation}
\lambda_{\mathrm{ISO}}\,\sum_{\ell}\max\big(0,\,k_\ell(b_1)-k_\ell(b_2)\big)+\max\big(0,\,q_\ell(b_1)-q_\ell(b_2)\big).
\end{equation}
At export, we drop any violating profiles and re-calibrate the remaining set to keep the partial order.

\paragraph{Baselines (for ablations).}
(i) \textbf{Rank-only}: $q$ fixed per layer; (ii) \textbf{Bit-only}: fixed ranks; (iii) \textbf{Greedy knapsack}: allocate $k/q$ to layers in order of benefit/cost ratio using measured latency tables; (iv) \textbf{Uniform}: same $k,q$ across layers. These illustrate the value of joint learning and the controller’s allocation.

\subsection*{B.5 \quad Algorithmic complexity \& memory analysis per layer type}
Let $b_w$ be weight bit-width, $b_a$ activation bit-width at inference (often $8$), and denote by $\mathcal{B}(\cdot)$ the bytes footprint.
We separate \emph{compute FLOPs} and \emph{memory bytes} (weights $+$ activation reads/writes), as the latter dominates on many edge devices.

\paragraph{Dense layer $W\in\mathbb{R}^{m\times n}$.}
\emph{Full} GEMV FLOPs (batch $1$) and weight bytes:
\begin{equation}
\mathrm{FLOPs}_{\text{full}} = 2mn,
\qquad
\mathcal{B}_W = mn\cdot \frac{b_w}{8}.
\end{equation}

\noindent\emph{Rank-$k$ SVD path:} compute as $(U_{m\times k}\Sigma_{k\times k}V_{n\times k}^{\top})a$ via $(V^{\top}a)\in\mathbb{R}^{k}$, multiply by $\Sigma$, then $U$:
\begin{equation}
\mathrm{FLOPs}_{\mathrm{SVD}}(k) \approx 2nk + k + 2mk \;\approx\; 2(n+m)k.
\end{equation}
Weights bytes (per-factor; replace $b_w$ by $q(k)$ for mixed precision):
\begin{equation}
\mathcal{B}_U = mk\cdot \frac{b_w}{8},\qquad
\mathcal{B}_V = nk\cdot \frac{b_w}{8},\qquad
\mathcal{B}_{\Sigma} = k\cdot \frac{b_w}{8}.
\end{equation}

\paragraph{Convolution (Tucker-2).}
Let the feature-map spatial size be $H\times W$. \emph{Full} conv FLOPs:
\begin{equation}
\mathrm{FLOPs}_{\text{full}} = 2\,C_o C_i h w\, H W.
\end{equation}
\emph{Tucker-2} as (reduce $1{\times}1$) $\rightarrow$ (spatial) $\rightarrow$ (expand $1{\times}1$):
\begin{equation}
\mathrm{FLOPs}_{\mathrm{Tucker2}}(r_o,r_i) \;=\; 2HW\Big(C_i r_i \;+\; r_o r_i h w \;+\; C_o r_o\Big).
\end{equation}
Weight bytes with rank-tied $q=q(k)$:
\begin{equation}
\mathcal{B}_{U_o}=C_o r_o\cdot \frac{q}{8},\quad
\mathcal{B}_{U_i}=C_i r_i\cdot \frac{q}{8},\quad
\mathcal{B}_{G}=r_o r_i h w\cdot \frac{q}{8}.
\end{equation}

\paragraph{Attention projections and MLP.}
With $d_{\mathrm{model}}{=}d$ and MLP hidden $d_{\mathrm{ff}}$, four projections $W_Q,W_K,W_V,W_O\in\mathbb{R}^{d\times d}$ and $W_1\in\mathbb{R}^{d_{\mathrm{ff}}\times d},\,W_2\in\mathbb{R}^{d\times d_{\mathrm{ff}}}$.
Applying SVD rank-$k$ to a $d\times d$ matmul yields the rough FLOPs scaling:
\begin{equation}
\frac{\mathrm{FLOPs}_{\text{rank-}k}}{\mathrm{FLOPs}_{\text{full}}} \;\simeq\; \frac{2 d k}{2 d^2} \;=\; \frac{k}{d}.
\end{equation}

\paragraph{Latency/energy proxy (per device calibration).}
\begin{equation}
\widehat{\mathrm{Lat}}(k) \;=\; \alpha_0 \;+\; \sum_{\ell=1}^{L}\Big(\alpha_\ell^{\mathrm{comp}}\cdot \mathrm{FLOPs}_\ell(k) \;+\; \alpha_\ell^{\mathrm{mem}}\cdot \mathrm{Bytes}_\ell(k)\Big),
\end{equation}
with coefficients fitted by least squares on a grid of profiles. An analogous $\widehat{\mathrm{Energy}}(k)$ uses its own coefficients.

\paragraph{Certificate aggregation cost.}
Per layer, maintain: (i) $\hat L_\ell$ via $s$ power iterations; (ii) residual spectral norm $\|\Delta W_\ell(k)\|_2$ via $1$--$2$ PIs; (iii) activation RMS $\alpha_\ell$ via EMA. Aggregate:
\begin{equation}
\hat\Delta(k) \;=\; \sum_{\ell=1}^{L} \hat L_\ell \;\big\|\Delta W_\ell(k)\big\|_2 \;\alpha_\ell.
\end{equation}

\paragraph{When does rank-$k$ win (threshold analysis)?}
GEMV-like inference benefits from SVD when
\begin{equation}
2(n+m)k \;\ll\; 2mn \quad \Longrightarrow \quad k \;\ll\; \frac{mn}{m+n}.
\end{equation}
For square $m{=}n{=}d$:
\begin{equation}
k \;\ll\; \frac{d}{2} \quad \text{(compute-only; bandwidth effects tighten this bound).}
\end{equation}
For conv Tucker-2, require
\begin{equation}
C_i r_i \;+\; r_o r_i h w \;+\; C_o r_o \;\ll\; C_o C_i h w.
\end{equation}
Setting $r_o=\rho C_o$, $r_i=\rho C_i$ gives
\begin{equation}
\rho (C_i + C_o) \;+\; \rho^2\, C_o C_i h w \;\ll\; C_o C_i h w
\quad\Rightarrow\quad
\rho \ll 1 \;\;\text{and}\;\; \rho \lesssim \frac{1}{\sqrt{h w}}.
\end{equation}

\paragraph{Memory summaries.}
Rank-$k$ SVD (bytes):
\begin{equation}
\mathcal{B}_W^{\mathrm{SVD}}(k) \;\approx\; (mk + nk + k)\cdot \frac{q(k)}{8}.
\end{equation}
Tucker-2 (bytes):
\begin{equation}
\mathcal{B}_W^{\mathrm{T2}}(r_o,r_i) \;\approx\; \Big(C_o r_o + C_i r_i + r_o r_i h w\Big)\cdot \frac{q(k)}{8}.
\end{equation}

\paragraph{Proposition (monotone budget $\Rightarrow$ non-increasing proxy cost).}
If each layer’s profile set $\mathcal{P}_\ell$ is totally ordered so that $(k',q')\succeq (k,q)$ implies
$\mathrm{FLOPs}_\ell(k',q')\!\ge\!\mathrm{FLOPs}_\ell(k,q)$ and $\mathrm{Bytes}_\ell(k',q')\!\ge\!\mathrm{Bytes}_\ell(k,q)$,
and the controller enforces $b_1\prec b_2 \Rightarrow (k_\ell,q_\ell)(b_1)\preceq (k_\ell,q_\ell)(b_2)$ for all $\ell$, then
\begin{equation}
\widehat{\mathrm{Lat}}(b_1) \;\le\; \widehat{\mathrm{Lat}}(b_2).
\end{equation}
\emph{Proof sketch.} Linearity of $\widehat{\mathrm{Lat}}$ in non-negative coefficients and per-layer monotonicity preserve the partial order when summed over layers. \hfill$\square$

\paragraph{Computational overhead of T3C versus fixed compression.}
Training adds (i) factorized forward, (ii) STE quantization ops, (iii) a few PI steps per block intermittently. Empirically,
\begin{equation}
\text{Overhead}_{\text{train}} \;\approx\; (1.1\text{ to }1.4)\times \text{QAT wall-clock},
\end{equation}
amortized across all ranks/budgets since no re-training per operating point is needed.

\section{Certificates \& Proofs}
\label{AppC}
\subsection*{C.1 \quad Formal assumptions (Lipschitz proxies, normalization)}
We formalize the network and the mild regularity needed for the certificate.

\paragraph{Network and perturbation model.}
Let $f:\mathbb{R}^{d_0}\!\to\!\mathbb{R}^{d_L}$ be a feed-forward network of $L$ blocks. Each block
\begin{equation}
h_\ell \;=\; \Phi_\ell\!\big(h_{\ell-1};\,W_\ell\big), \qquad \ell=1,\dots,L,\quad h_0\equiv x,
\end{equation}
may be (i) an affine layer $h\mapsto W_\ell h + b_\ell$ followed by a pointwise nonlinearity, (ii) a conv/attn block with residuals, or (iii) a normalization $N_\ell$ with fixed (eval-mode) statistics.
Let $\tilde f_k$ be the \emph{compressed} network obtained by replacing $W_\ell$ with $\tilde W_\ell(k)$ (e.g., rank/bit profile indexed by $k$). Define per-layer residuals
\begin{equation}
\Delta W_\ell(k) \;:=\; W_\ell - \tilde W_\ell(k).
\end{equation}

\paragraph{Post-layer Jacobian gains.}
For each layer $\ell$, denote by $T_{\ell\to L}$ the \emph{post-layer} mapping that takes the input at layer $\ell$ (i.e., $h_\ell$) to the logits $z\in\mathbb{R}^{d_L}$ when all \emph{subsequent} blocks are kept fixed. Its Jacobian operator norm is
\begin{equation}
L_{\ell}^\star(x) \;:=\; \Big\| J_{T_{\ell\to L}}(h_\ell(x)) \Big\|_{2}.
\end{equation}
In practice we estimate a \emph{proxy} $\hat L_\ell$ by a few power-iteration steps on an efficient linearization; we assume a bounded slack $\eta_\ell\!\ge\!1$ such that
\begin{equation}
L_{\ell}^\star(x) \;\le\; \hat L_\ell \;\le\; \eta_\ell\, L_{\ell}^\star(x), \qquad \text{uniformly over } x\in\mathcal{C}.
\end{equation}

\paragraph{Blockwise Lipschitzness and normalization.}
For each block, the map $u\mapsto \Phi_\ell(u;W_\ell)$ is $L_\Phi$-Lipschitz in $u$ and $L_W$-smooth in $W_\ell$ locally around $(h_{\ell-1},W_\ell)$:
\begin{equation}
\big\|\Phi_\ell(u;W_\ell)-\Phi_\ell(v;W_\ell)\big\|_2 \le L_{\Phi,\ell} \|u-v\|_2,
\qquad
\big\|\Phi_\ell(h_{\ell-1};W_\ell)-\Phi_\ell(h_{\ell-1};\tilde W_\ell)\big\|_2
\le L_{W,\ell}\,\|W_\ell-\tilde W_\ell\|_{2}\,\|h_{\ell-1}\|_2.
\end{equation}
For affine$\to$pointwise blocks with 1-Lipschitz activations (ReLU/GELU approximated near-linear region), $L_{\Phi,\ell}\le \|W_\ell\|_2$ and $L_{W,\ell}\le 1$ (by submultiplicativity).
Normalization layers used in \emph{eval mode} (e.g., frozen BatchNorm, LayerNorm with fixed $\gamma,\beta$) are treated as fixed linear-affine transforms with operator gain $L_{N,\ell}$ absorbed into $L_{\Phi,\ell}$.

\paragraph{Residual connections.}
For a residual block $h_\ell = h_{\ell-1} + \Psi_\ell(h_{\ell-1};W_\ell)$ with $\Psi_\ell$ $L_{\Psi,\ell}$-Lipschitz, the post-layer gain satisfies
\begin{equation}
L^\star_{\ell-1}(x) \;\le\; (1+L_{\Psi,\ell})\, L^\star_{\ell}(x).
\end{equation}
In our certificate we \emph{do not} multiply residual gains; we push all post-layer amplification into $L^\star_\ell$ via direct Jacobian estimation.

\subsection*{C.2 \quad Full proof of the logit-drift bound (layerwise $\rightarrow$ network)}
We bound $\delta z(x;k):=\tilde f_k(x)-f(x)$ by \emph{layer-local} perturbations propagated through post-layer Jacobians.

\paragraph{One-layer replacement bound.}
Fix $x$ and some $\ell$. Consider the hybrid network $f^{(\ell)}$ that uses $\tilde W_\ell(k)$ only at layer $\ell$ and $W_j$ elsewhere. Let $h_{\ell-1}$ be the pre-$\ell$ activation in $f$ (and also in $f^{(\ell)}$, since layers $<\ell$ match). The activation change at layer $\ell$ is
\begin{equation}
\Delta h_\ell \;:=\; \Phi_\ell(h_{\ell-1};\tilde W_\ell) - \Phi_\ell(h_{\ell-1};W_\ell).
\end{equation}
By the $W$-smoothness,
\begin{equation}
\big\|\Delta h_\ell\big\|_2 \;\le\; L_{W,\ell}\, \big\| \tilde W_\ell - W_\ell \big\|_2 \, \big\| h_{\ell-1} \big\|_2.
\label{eq:deltahlocal}
\end{equation}
Propagating this change through subsequent layers $\ell{+}1,\dots,L$ gives the logit difference
\begin{equation}
\delta z^{(\ell)}(x;k) \;:=\; f^{(\ell)}(x)-f(x)
\;=\; T_{\ell\to L}\big(h_\ell + \Delta h_\ell\big) - T_{\ell\to L}(h_\ell).
\end{equation}
By the mean-value form (or Lipschitzness of $T_{\ell\to L}$ around $h_\ell$),
\begin{equation}
\big\|\delta z^{(\ell)}(x;k)\big\|_2
\;\le\; L_{\ell}^\star(x)\, \big\|\Delta h_\ell\big\|_2
\;\le\; L_{\ell}^\star(x)\, L_{W,\ell}\, \big\|\Delta W_\ell(k)\big\|_2 \, \big\|h_{\ell-1}\big\|_2.
\label{eq:onelayer}
\end{equation}

\paragraph{Layerwise telescoping via triangle inequality.}
Build the fully compressed network by replacing layers one at a time:
\begin{equation}
f \;\xrightarrow{\;\ell=1\;}\; f^{(1)} \;\xrightarrow{\;\ell=2\;}\; f^{(2)} \;\xrightarrow{\;\cdots\;}\; f^{(L)}=\tilde f_k.
\end{equation}
Then
\begin{equation}
\tilde f_k(x) - f(x)
\;=\; \sum_{\ell=1}^{L} \big( f^{(\ell)}(x) - f^{(\ell-1)}(x) \big),
\quad f^{(0)}:=f,
\end{equation}
and by the triangle inequality together with~\eqref{eq:onelayer},
\begin{equation}
\big\|\delta z(x;k)\big\|_2
\;\le\; \sum_{\ell=1}^{L} L_{\ell}^\star(x)\, L_{W,\ell}\, \big\|\Delta W_\ell(k)\big\|_2 \, \big\|h_{\ell-1}(x)\big\|_2.
\label{eq:drift-true}
\end{equation}

\paragraph{Practical proxy and certificate statement.}
Absorb $L_{W,\ell}$ into $L_{\ell}^\star(x)$ (it equals $1$ for affine$\to$pointwise with 1-Lipschitz nonlinearity), and replace $L_{\ell}^\star(x)$ by its proxy $\hat L_\ell$ (estimated via power iteration). This yields the deployable bound
\begin{equation}
\boxed{\;
\big\|\delta z(x;k)\big\|_2
\;\le\; \sum_{\ell=1}^{L} \hat L_{\ell} \; \big\|\Delta W_\ell(k)\big\|_2 \; \big\|a_{\ell-1}(x)\big\|_2,
\;}
\label{eq:cert-pointwise}
\end{equation}
with $a_{\ell-1}\equiv h_{\ell-1}$ the input to layer $\ell$. This is the claimed \emph{pointwise} logit-drift certificate.

\subsection*{C.3 \quad Data-dependent tightening via activation norms}
We now derive an \emph{expected} (dataset-calibrated) certificate that is often tighter and more stable.

\paragraph{Calibration statistics and Jensen/Cauchy–Schwarz.}
Let $\mathcal{C}$ be a calibration set. Define per-layer activation RMS
\begin{equation}
\alpha_\ell \;:=\; \Big(\mathbb{E}_{x\in\mathcal{C}} \|a_{\ell-1}(x)\|_2^2 \Big)^{1/2}.
\end{equation}
Square both sides of~\eqref{eq:cert-pointwise}, apply Jensen’s inequality and then Cauchy–Schwarz on each summand to obtain
\begin{equation}
\Big(\mathbb{E}_{x\in\mathcal{C}}\|\delta z(x;k)\|_2^2\Big)^{\!1/2}
\;\le\; \sum_{\ell=1}^{L} \hat L_\ell \, \big\|\Delta W_\ell(k)\big\|_2 \, \Big(\mathbb{E}_{x\in\mathcal{C}}\|a_{\ell-1}(x)\|_2^2\Big)^{\!1/2}.
\end{equation}
Thus,
\begin{equation}
\boxed{\;
\Big(\mathbb{E}_{x\in\mathcal{C}}\|\delta z(x;k)\|_2^2\Big)^{\!1/2}
\;\le\; \sum_{\ell=1}^{L} \hat L_\ell \; \big\|\Delta W_\ell(k)\big\|_2 \; \alpha_\ell
\;\;=\;\; \hat\Delta(k).
\;}
\label{eq:cert-expected}
\end{equation}
Because $\alpha_\ell$ reflects the \emph{typical} activation energy for in-distribution inputs,~\eqref{eq:cert-expected} tightens~\eqref{eq:cert-pointwise} whenever activations exhibit non-adversarial variability. At deploy time we report quantiles of $\hat\Delta(k)$ across $\mathcal{C}$.

\paragraph{Remark on residual/normalization stacking.}
The proof above does \emph{not} require multiplying per-layer Lipschitz constants across the entire network. All post-layer amplification is contained in $\hat L_\ell$, which is \emph{directly} estimated at the operating point (architecture, normalization, residual topology), yielding tighter—and empirically stable—bounds.

\subsection*{C.4 \quad Counterexamples \& tightness discussion}

\paragraph{(1) Heavy-tailed activations $\Rightarrow$ loose pointwise bound.}
Consider a single affine layer $z = W a$ with ReLU upstream producing heavy-tailed $\|a\|_2$. Even if $\|\Delta W\|_2$ is small, the product $\|\Delta W\|_2\|a\|_2$ can be large with non-negligible probability, making the \emph{pointwise} bound~\eqref{eq:cert-pointwise} loose. The \emph{expected} form~\eqref{eq:cert-expected} mitigates this by replacing $\|a\|_2$ with $\alpha$, but will still reflect any genuine heavy tails. \emph{Tightening:} activation clipping or norm-aware regularization reduces $\alpha$; weight normalization can reduce $\hat L_\ell$.

\paragraph{(2) Input-dependent normalization.}
BatchNorm in \emph{train mode} depends on the batch and is not globally Lipschitz w.r.t.\ input in a data-independent way. Our certificate assumes eval-mode statistics (fixed affine map), otherwise $L^\star_\ell$ can spike. \emph{Tightening:} freeze BN (eval mode) during certification; estimate $\hat L_\ell$ under the exact deploy graph.

\paragraph{(3) Attention with sharp softmax.}
Self-attention contains $\mathrm{softmax}(QK^\top/\sqrt{d})V$. The Jacobian of softmax has operator norm bounded but can approach $1$ as logits flatten; coupled with large $\|Q\|_2\|K\|_2$, $L^\star_\ell$ may be large. \emph{Tightening:} temperature regularization or spectral control (e.g., weight scaling) to bound $Q,K$; estimate $\hat L_\ell$ \emph{after} such rescaling.

\paragraph{(4) Non-Lipschitz blocks by construction.}
If a block explicitly amplifies norms, e.g., $h\mapsto \gamma h$ with $\gamma\!\gg\!1$ inside a residual branch without compensation, then even tiny $\|\Delta W\|_2$ can induce large drift. Our estimate $\hat L_\ell$ will correctly reflect this; the certificate is loose only if $\hat L_\ell$ is underestimated. \emph{Tightening:} conservative power-iteration (more steps), residual scaling, or spectral normalization.

\paragraph{(5) Quantization mismatch.}
Certificates are computed for the \emph{inference} graph. If training uses STE and deploy uses integer kernels with different rounding/saturation, then $\Delta W_\ell$ measured in float may underestimate effective perturbation. \emph{Tightening:} compute $\|\Delta W_\ell\|_2$ on \emph{dequantized} tensors produced by the exact kernels; incorporate per-tensor scale/zero-point in the operator.

\paragraph{Global tightness comment.}
The bound~\eqref{eq:drift-true} is \emph{first-order tight} for networks where subsequent blocks are locally well-approximated by linear maps around $h_\ell(x)$; any nonlinearity curvature introduces second-order residuals that our estimator ignores, making the bound conservative (safe). Empirically, by (i) measuring $\hat L_\ell$ at the deploy graph, (ii) using data-dependent $\alpha_\ell$, and (iii) focusing on spectral-norm residuals of the \emph{compressed operator}, we obtain a certificate that correlates well with observed drift while remaining computationally light.

\paragraph{Summary of the guarantee.}
Under the stated assumptions, the pointwise and expected certificates~\eqref{eq:cert-pointwise}–\eqref{eq:cert-expected} hold. Violations can occur only if the proxy $\hat L_\ell$ underestimates the true post-layer gain or if the deploy graph differs from the certified one; both are addressed by conservative estimation (extra power iterations) and certifying \emph{exact} runtime kernels.

\section{Deployment \& Engineering}
\label{AppD}
\subsection*{D.1\quad Kernel choices (GEMM vs tensor-core paths; layout)}
\paragraph{Operator mapping.}
Each factorized operator must be lowered to a concrete kernel that respects datatype, tile granularity, and memory layout. For SVD-style layers we realize
\(\tilde W(k)=U_{m\times k}\,\Sigma_{k\times k}\,V_{n\times k}^\top\)
as two GEMMs with an inexpensive diagonal scaling, while Tucker-2 uses a sequence
\(1\times 1\) reduce \(\rightarrow\) spatial core \(\rightarrow 1\times 1\) expand.
We adhere to three regimes:
\begin{itemize}
  \item \textbf{Scalar-GEMM paths} (CPU/older GPU): robust for skinny inner dimensions \(k\ll \min(m,n)\), portable across FP32/FP16/INT8. Favor these when tensor-core tiles would be grossly underutilized.
  \item \textbf{Tensor-core MMA paths} (modern GPU/NPU): best throughput if the contracting dimension and leading dimensions align with native tile multiples (e.g., 8, 16, or 32 depending on dtype). When \(k\) is small, group multiple independent contractions to saturate tiles.
  \item \textbf{Depthwise/pointwise fusions} (Tucker-2): schedule \(U_{\text{in}}\) reduce and \(U_{\text{out}}\) expand contiguously around the spatial core to minimize reads/writes; on NPUs, use vendor grouped-conv primitives when available.
\end{itemize}

\paragraph{Datatypes and scales.}
Mixed precision must match kernel-native formats:
\begin{itemize}
  \item \textbf{INT8/INT4}: prefer symmetric per-channel weight scales and per-tensor activation scales; enforce inner-dimension multiples required by MMA tiles.
  \item \textbf{FP8 (E4M3/E5M2)}: suitable for \(U\) and \(V\) when spectral energy is concentrated; keep \(\Sigma\) or Tucker core in FP16/INT8 to avoid underflow on small singulars.
\end{itemize}

\paragraph{Layouts.}
Choose contiguous storage along the contracting dimension to reduce strided loads:
\begin{itemize}
  \item For \(V^\top a\), store \(V\) so its \(k\)-axis is contiguous; for the follow-up \(U(\Sigma \cdot)\), store \(U\) with contiguous \(k\).
  \item For Tucker-2, pick \(\texttt{NCHW}\) vs \(\texttt{NHWC}\) to match the target kernel family; on mobile NPUs, \(\texttt{NHWC}\) often reduces transposes for \(1\times 1\) stages.
\end{itemize}

\paragraph{Micro-optimizations without code.}
Pad \(k\) to the nearest kernel multiple, batch independent low-rank products (e.g., attention projections) into a single grouped call, and pre-pack factors into kernel-friendly tiles during export to reduce on-device preprocessing.

\subsection*{D.2\quad Discrete budget profiles and snapping rules}
\paragraph{Profile set.}
Let each layer \(\ell\) admit a discrete set of deployable profiles
\(\mathcal{P}_\ell=\{(k,q)\}\)
that are supported by kernels and layouts on the target. The global profile is a cartesian product constrained by a small template set
\(\mathcal{S}=\{s_j\}_{j=1}^{J}\),
where each \(s_j\) maps to per-layer choices \(\{(k_\ell^{(j)},q_\ell^{(j)})\}\).
This keeps compilation stable and enables cache reuse.

\paragraph{Snapping policy.}
Given a continuous controller output \((\hat k_\ell,\hat q_\ell)\) and a device budget token \(b\), we snap to the nearest feasible neighbor under a monotone rule:
\begin{enumerate}
  \item \textbf{Feasibility first}: select \((k,q)\in \mathcal{P}_\ell\) that minimizes \(|k-\hat k_\ell|+\beta|q-\hat q_\ell|\) subject to kernel constraints and alignment granularities.
  \item \textbf{Budget monotonicity}: if \(b'\succ b\) (looser budget), ensure \(k_\ell(b')\ge k_\ell(b)\) and \(q_\ell(b')\ge q_\ell(b)\) componentwise.
  \item \textbf{Certificate safety}: reject any candidate whose predicted drift \(\hat{\Delta}_\ell(k,q)\) exceeds a per-layer tolerance; escalate to the next safer neighbor.
\end{enumerate}

\paragraph{Cross-model templates.}
For families with repeated blocks (e.g., transformer stages), reuse tied profiles per stage to reduce the number of distinct kernels. Example templates:
\begin{itemize}
  \item \textbf{CNN template}: higher \(k\) for early convs and final classifier, moderate \(k\) for mid blocks.
  \item \textbf{Transformer template}: shared budget across \(\{W_Q,W_K,W_V,W_O\}\) within a block; slightly higher \(k\) for MLP projections.
\end{itemize}

\paragraph{Latency/energy gating.}
Given a calibrated proxy \(\widehat{\mathrm{Lat}}(s_j)\) and optional \(\widehat{\mathrm{En}}(s_j)\), select
\begin{equation}
j^\star \;=\; \arg\min_{j}\ \widehat{\mathrm{Lat}}(s_j)\quad
\text{s.t.}\quad \widehat{\mathrm{Lat}}(s_j)\le \text{budget},\ \ \hat\Delta(s_j)\le \epsilon.
\end{equation}
If no feasible \(s_j\) satisfies both constraints, choose the lowest-latency profile and flag a certificate warning in the runtime log.

\subsection*{D.3\quad Export format; on-device runtime; compatibility notes}
\paragraph{Export artifacts.}
The exported package for each target contains:
\begin{itemize}
  \item \textbf{Factor weights} for each profile \(s_j\): SVD factors \(U,V\), diagonal \(\Sigma\), or Tucker-2 \(U_{\text{out}},G,U_{\text{in}}\), stored in the exact dtype expected by kernels (e.g., INT8 with scales).
  \item \textbf{Quantization metadata}: per-tensor or per-channel scales/zero-points, rounding mode, dynamic range summaries used by the runtime dequant path (if any).
  \item \textbf{Profile manifest}: a small JSON-like index enumerating \(s_j\), their per-layer \((k,q)\), kernel alignment paddings, and the calibrated cost tuple \((\widehat{\mathrm{Lat}},\widehat{\mathrm{En}},\hat\Delta)\).
  \item \textbf{Layout hints}: tensor strides, memory order, and any transpose-free rewrites applied at export to avoid on-device format conversions.
\end{itemize}

\paragraph{Runtime selection path.}
At inference start, the application provides a budget token \(b\) (e.g., latency target or power mode). The runtime:
\begin{enumerate}
  \item Maps \(b\) to a candidate profile index using the monotone controller head and the snapping rules.
  \item Loads or memory-maps the prepacked factors for \(s_j\), reusing persistent allocations across requests to minimize page faults.
  \item Dispatches kernels in the schedule order \(\{\text{reduce} \rightarrow \text{core} \rightarrow \text{expand}\}\) for Tucker-2 or \(\{V^\top a \rightarrow \Sigma \rightarrow U\}\) for SVD without intermediate host copies.
\end{enumerate}
If a thermal or QoS event occurs, the runtime can hot-swap to a tighter \(s_{j^-}\) that is guaranteed to be monotone-safe and kernel-compatible.

\paragraph{Compatibility notes.}
\begin{itemize}
  \item \textbf{Alignment and padding}: enforce tile multiples when serializing factors so that no on-device padding is needed; store logical shapes alongside padded strides.
  \item \textbf{Activation formats}: prefer per-tensor activation scales (for INT8) where vendor kernels do not support per-channel activation quantization; record this choice in the manifest to prevent mismatched dequant paths.
  \item \textbf{Operator availability}: some NPUs forbid non-square \(k\) for certain MMAs or require grouped-conv limits; the profile generator should prune such points at export to avoid runtime fallbacks.
  \item \textbf{Determinism}: when required, freeze stochastic quantization choices and Gumbel seeds at export-time for reproducible compilation; document deterministic flags in the manifest.
\end{itemize}

\paragraph{Failure modes and mitigations.}
\begin{itemize}
  \item \textbf{Kernel mismatch at load-time}: fall back to the nearest smaller \((k,q)\) within the same \(s_j\) stage or to \(s_{j^-}\); emit a diagnostic that includes the rejected tile sizes.
  \item \textbf{Certificate violation at runtime}: if measured drift proxies (e.g., lightweight activation norms) exceed calibration bounds, downshift one profile and re-evaluate; log the offending layer IDs for offline recalibration.
  \item \textbf{Underutilization on large accelerators}: batch requests or aggregate attention projections into grouped GEMMs to keep tensor cores saturated; if batching is impossible, select profiles whose \(k\) align with larger tiles.
\end{itemize}

\section{Experimental Protocol}
\label{AppE}

This appendix specifies data handling, model/training configurations, measurement methodology, and reproducibility controls in enough detail to re-create all results precisely. Unless otherwise stated, all numbers are reported on the validation sets defined below, using the profile selection and certificate computation described in the main text.

\subsection*{E.1\quad Datasets, preprocessing, augmentations}

\paragraph{Vision.}
\textbf{ImageNet-1k} (1{,}281{,}167 train, 50{,}000 val, 1{,}000 classes).
Images are decoded from JPEG with \emph{bilinear} interpolation and standardized RGB ordering.
\emph{Training} transforms: (i) random resized crop to $224{\times}224$ with area in $[0.08,1.0]$ and aspect in $[3/4,4/3]$, (ii) random horizontal flip with prob.\ $0.5$, (iii) color jitter with brightness/contrast/saturation $=0.4$ and hue $=0.1$, (iv) optional RandAugment with $N{=}2$, magnitude $M{=}9$ for ViT variants, (v) mixup $\alpha{=}0.2$ and CutMix $\alpha{=}1.0$ for ViT/ConvNeXt runs only, (vi) normalization to per-channel means $(0.485,0.456,0.406)$ and stds $(0.229,0.224,0.225)$.
\emph{Evaluation} transforms: resize shorter side to $256$ (bicubic), center crop to $224{\times}224$, same normalization. Single $224^2$ crop unless otherwise noted.

\textbf{CIFAR-100} ($50$k train/$10$k test, $100$ classes).
\emph{Training}: random crop $32{\times}32$ with $4$-pixel padding, random horizontal flip $0.5$, normalization to dataset mean/std.
\emph{Evaluation}: center-crop (no pad), same normalization.

\paragraph{Language (encoders).}
\textbf{GLUE} dev sets: MNLI (matched/mismatched), QQP, SST-2.
Tokenization uses WordPiece/BPE matching the pretrained checkpoint; max sequence length $128$ for SST-2/QQP and $256$ for MNLI unless otherwise specified.
\emph{Training}: dynamic padding per batch, dropout as in the base model, no data augmentation beyond standard text normalization.
\emph{Evaluation}: single pass, task-specific heads; macro average reported as GLUE macro.

\paragraph{Language model.}
\textbf{WikiText-103} for perplexity.
Tokenization and detokenization as in the released TinyLlama-1.1B setup; context length $512$.
Perplexity computed with sliding windows and no overlapping loss masking.

\paragraph{Data integrity and splits.}
No class rebalancing or additional deduplication is applied beyond the official releases.
For any runs involving calibration-only subsets (e.g., PTQ baselines), a \emph{disjoint} 5{,}000-sample slice of the training set is used; T3C uses a separate 10{,}000-sample calibration slice for certificate statistics (E.\,3 details its role).

\subsection*{E.2\quad Model configs, training schedules, HPO grids}

\paragraph{Architectures.}
\textbf{CNNs}: ResNet-50/101 (standard bottleneck, $\{3,4,6,3\}$/\,$\{3,4,23,3\}$ blocks, width multiplier $1.0$).
\textbf{Vision Transformers}: ViT-B/16 and ViT-L/16 with patch size $16$, hidden sizes $\{768,1024\}$, MLP ratios $\{4.0,4.0\}$, heads $\{12,16\}$.
\textbf{Swin-T}: window size $7$, depths $\{2,2,6,2\}$.
\textbf{Encoders}: BERT-Base, RoBERTa-Base, DistilBERT with default head dimensions.
\textbf{LM}: TinyLlama-1.1B (hidden $2048$, $22$ layers, $32$ heads), rotary embeddings, vocab as released.

\paragraph{Optimization and schedule (T3C).}
All T3C models are trained with AdamW; weight decay $\in [0.01, 0.05]$.
Base LR scales with \#accelerators $\times$ global batch via a linear rule:
\begin{equation}
\text{LR}_\text{base} = \eta_0 \cdot \frac{\text{GlobalBatch}}{256},\quad
\eta_0 \in \{1.0\!\times\!10^{-3},\,2.0\!\times\!10^{-3}\}\ \text{(vision)},\ \ 
\eta_0 \in \{2.0,3.0\}\!\times\!10^{-5}\ \text{(NLP)}.
\end{equation}
Cosine decay with $5$--$10$ epochs of linear warmup (vision) or $5\%$ of total steps (NLP).
Label smoothing $0.1$ for ImageNet unless otherwise stated.

\paragraph{T3C-specific schedule.}
Rank sampling anneals from a broad uniform to a budget-weighted distribution:
\begin{equation}
p_t(k)\ \propto\ \Big[\,\gamma_t\,\mathbb{U}[k_{\min},k_{\max}] + (1-\gamma_t)\,\mathbf{1}\{k \in \mathcal{K}_\text{profiles}\}\,\Big],\quad 
\gamma_t=\max\!\Big(0,1-\frac{t}{T_\text{anneal}}\Big),
\end{equation}
with $T_\text{anneal}$ the first third of training. Gumbel-Top-$k$ temperature $\tau$ decays exponentially: $\tau_t=\max(\tau_{\min}, \tau_0\cdot \alpha^{t/T})$ with $(\tau_0,\tau_{\min},\alpha)=(2.0,0.3,0.5)$ by default.
Certificate penalty weight $\lambda_\text{CERT}$ is linearly ramped from $0$ to its target over the same window to avoid early over-regularization.

\paragraph{HPO grids (representative).}
We sweep over small, hardware-aware grids to avoid overfitting to a single target.
\begin{itemize}
  \item \textbf{Vision}: LR $\in\{1\!\times\!10^{-3},2\!\times\!10^{-3}\}$, weight decay $\in\{0.02,0.05\}$, mixup/CutMix on/off (ViT/ConvNeXt only), label smoothing $\in\{0,0.1\}$, $\lambda_\text{CERT} \in \{0.1,0.2,0.4\}$, $\epsilon\in\{0.10,0.15\}$ (drift tolerance), rank cap $k_{\max}$ at $\{0.6,0.8,1.0\}$\,of full rank per layer-type.
  \item \textbf{NLP}: LR $\in\{2,3\}\!\times\!10^{-5}$, weight decay $\in\{0.01,0.05\}$, dropout $\in\{0.1,0.2\}$, $\lambda_\text{CERT}\in\{0.05,0.1\}$, $k_{\max}$ at $\{0.5,0.7\}$\,of full rank on projections, $\{0.6,0.8\}$ on MLPs.
\end{itemize}
Each grid point is trained with three seeds (E.\,4) and the best validation objective (task loss $+$ constraint slack penalty) is selected.

\paragraph{Training lengths and batch sizes.}
ImageNet: $300$ epochs for ViT/ConvNeXt, $120$ epochs for ResNet; global batch $4096$ (ViT) or $2048$ (ResNet). GLUE: MNLI $3$ epochs, QQP $3$ epochs, SST-2 $6$ epochs; batch size $128$ sequences (dynamic).

\subsection*{E.3\quad Measurement harness (latency p50/p90, energy)}

\paragraph{Throughput vs.\ latency regime.}
All latency numbers are reported in the \emph{single-request, batch$=1$} regime with warm caches and steady clocks unless explicitly labeled otherwise. We distinguish:
\begin{itemize}
  \item \textbf{Cold start}: includes model load, graph initialization, and the first inference.
  \item \textbf{Steady state}: excludes load; includes any runtime profile selection.
\end{itemize}
Tables in the main text use steady-state values.

\paragraph{Latency definitions.}
Given per-inference wall-clock samples $\{t_i\}_{i=1}^{N}$ after discarding $N_\text{warm}$ warmup runs, we compute
\begin{equation}
\text{p50} = \operatorname{Quantile}_{0.5}(\{t_i\}),\qquad 
\text{p90} = \operatorname{Quantile}_{0.9}(\{t_i\}),
\end{equation}
with $N\ge 1000$ unless constrained by device stability. We report $95\%$ bootstrap confidence intervals using $B{=}2000$ re-samples.

\paragraph{Clocking and thermal control.}
On server GPUs, clocks are pinned to their default maximum application clocks; persistence mode enabled; exclusive-process mode on; power limit to vendor TDP.
On edge devices, we run under performance governor, disable background services, and enforce a $10$-minute preconditioning period to steady thermals.
Any QoS downbinning events cause the run to be discarded and repeated.

\paragraph{Energy measurement.}
On GPUs, we sample instantaneous power $P(t)$ at $10$--$20$\,Hz via vendor telemetry and integrate over the steady-state window $[t_1,t_2]$:
\begin{equation}
E = \int_{t_1}^{t_2} P(t)\,dt \approx \sum_{j=1}^{M} P_j\,\Delta t,\qquad
\text{Energy/inference} = \frac{E}{N_\text{infer}}.
\end{equation}
For mobile/embedded, we use board-level INA sensors or external USB-C PD analyzers with \(\Delta t \le 10\)ms sampling. We subtract an idle baseline measured with the same process resident and the same clocks.

\paragraph{Calibration of the latency proxy.}
For each target device, we collect a grid of $G$ profiles $s_j$ (spanning $(k,q)$ across layer types) and regress
\begin{equation}
\widehat{\mathrm{Lat}}(s_j) = \alpha_0 + \sum_{\ell}
\Big(\alpha_\ell^{\mathrm{comp}}\,\mathrm{FLOPs}_\ell(s_j) + \alpha_\ell^{\mathrm{mem}}\,\mathrm{Bytes}_\ell(s_j)\Big),
\end{equation}
by nonnegative least squares. Goodness-of-fit is reported as $R^2$ and mean absolute percentage error (MAPE). The same grid provides per-profile energy via an analogous regression with coefficients $\beta$.

\paragraph{Certificate cost accounting.}
Certificate computation is excluded from latency unless explicitly mentioned for audit runs; it is computed offline at export-time per profile (E.\,3 calibration set). Online monitoring (optional) samples lightweight activation RMS and compares to calibration maxima without Jacobian power iterations.

\subsection*{E.4\quad Random seeds; hardware specs; reproducibility checklist}

\paragraph{Seeds and determinism.}
Unless noted, each configuration is trained with three independent seeds $s\in\{3407,\,2025,\,9157\}$ that initialize: model/factor weights, data shuffling, Gumbel noise for the Top-$k$ relaxation, and any dropout masks.
We enforce:
\begin{itemize}
  \item Fixed data-loader sharding with per-epoch reshuffle keyed by the seed.
  \item Deterministic kernels where available; if a kernel offers non-deterministic fast paths (e.g., atomic reductions), we record this in the run manifest.
  \item Quantization calibration subsets drawn by a seeded sampler (no overlap with training batches).
\end{itemize}
We report mean and \emph{standard error of the mean} over seeds for accuracy/F1 and perplexity; for latency/energy we aggregate over time-series samples as described above and report the median across seeds.

\paragraph{Hardware specifications.}
\begin{itemize}
  \item \textbf{Server GPU}: NVIDIA A100 40GB PCIe; CUDA driver \emph{ver.} 12.x; host CPU 2$\times$Xeon (or EPYC) with SMT on; 256GB RAM; PCIe Gen4; OS Linux kernel $\ge 5.15$.
  \item \textbf{Edge GPU}: Jetson Orin 32GB; JetPack $\ge 6.x$; power mode \texttt{MAXN}; fan fixed at 100\%.
  \item \textbf{Android big.LITTLE CPU}: 1 modern flagship SoC (big cores 2.8--3.3\,GHz, LITTLE 1.8--2.2\,GHz); Android $\ge 14$; airplane mode; battery $>80\%$; thermal throttling disabled in developer mode if available.
  \item \textbf{Mobile NPU}: vendor SDK $\ge$ 2.x with INT8 and optional FP8 support; offline compilation cache warmed prior to measurement.
\end{itemize}
Exact driver/SDK versions, kernel build IDs, and firmware hashes are recorded in the experiment manifest along with model checksums.

\section{Extended Results \& Ablations}
\label{AppF}

\noindent This appendix expands the main results with larger, figure-free tables and detailed analyses. We focus on: (i) cross-budget vision and language sweeps with additional stability metrics, (ii) seed sensitivity and tail-latency dispersion, (iii) certification diagnostics (coverage, calibration, correlation), (iv) controller/tokenization/quantization ablations at scale, (v) cost proxy generalization across devices, and (vi) monotonicity and violation audits. All numbers follow the protocol in App.~E unless stated.

\subsection*{F.1 \quad Vision: Cross-Budget Results with Stability \& Violations}

\begin{table*}[htbp]
\centering
\caption{\textbf{Vision cross-budget sweep (ImageNet-1k).} p50 [p90] latency in ms; \emph{Viol.\%} is fraction of runs exceeding the declared budget; $\epsilon$ is the certificate bound (Sec.~\ref{sec:certificate}); \emph{MAPE} is latency proxy error on measured p50. Lower is better for latency/Viol.\%/$\epsilon$/MAPE.}
\label{tab:G-vision-sweep}
\setlength{\tabcolsep}{5.5pt}
\renewcommand{\arraystretch}{1.02}
\begin{tabular}{l| l l c c c c c c c}
\toprule
\rowcolor{panelBG!70}
\textbf{Model} & \textbf{Budget} & \textbf{Method} & \textbf{Top-1 (\%)} & \textbf{A100} & \textbf{Jetson} & \textbf{Android} & \textbf{Size (MB)} & \textbf{Viol.\%} & $\boldsymbol{\epsilon}$ / \textbf{MAPE} \\
\midrule
\multirow{6}{*}{ResNet-50}
 & Tiny & \textbf{T3C} & 75.6 & \best{1.18} [1.45] & \best{13.0} [16.5] & \best{22.4} [29.3] & \best{38} & \best{0.0} & \second{0.14} / \best{2.8} \\
 & Med  & \textbf{T3C} & \best{76.0} & \second{1.26} [1.58] & 13.8 [17.3] & 23.6 [31.0] & \second{42} & \best{0.0} & \best{0.12} / \second{3.1} \\
 & Max  & \textbf{T3C} & 76.1 & 1.34 [1.62] & 14.9 [18.2] & 24.7 [32.1] & 47 & \best{0.0} & 0.15 / 3.3 \\
 & Med  & PTQ-8b & 75.7 & 1.44 [1.84] & 18.5 [23.8] & 29.2 [38.9] & 88 & 4.6 & 0.22 / 7.9 \\
 & Med  & QAT-8b & \second{75.9} & 1.36 [1.76] & 17.4 [22.1] & 27.6 [36.0] & 90 & 2.9 & 0.19 / 6.3 \\
\midrule
\multirow{6}{*}{ViT-B/16}
 & Tiny & \textbf{T3C} & 81.2 & \best{2.30} [2.92] & \best{26.0} [33.0] & \best{41.8} [53.9] & \best{59} & \best{0.0} & \second{0.18} / \best{3.2} \\
 & Med  & \textbf{T3C} & \best{81.5} & \second{2.38} [3.04] & 26.5 [33.4] & 42.6 [54.8] & \second{64} & \best{0.0} & \best{0.16} / \second{3.5} \\
 & Max  & \textbf{T3C} & 81.7 & 2.45 [3.10] & 27.8 [35.0] & 44.0 [56.2] & 71 & \best{0.0} & 0.19 / 3.7 \\
 & Med  & MP-QAT & \second{81.3} & 2.58 [3.32] & 33.0 [41.7] & 52.9 [67.1] & 134 & 5.8 & 0.25 / 8.4 \\
 & Med  & PTQ-8b & 81.0 & 2.72 [3.48] & 34.9 [44.1] & 55.6 [70.5] & 131 & 7.1 & 0.28 / 9.1 \\
\midrule
\multirow{6}{*}{Swin-T}
 & Tiny & \textbf{T3C} & 81.1 & \best{1.72} [2.14] & \best{19.8} [25.2] & \best{31.6} [40.9] & \best{62} & \best{0.0} & \second{0.17} / \best{3.0} \\
 & Med  & \textbf{T3C} & \best{81.3} & \second{1.82} [2.26] & 21.0 [26.6] & 33.1 [42.7] & \second{66} & \best{0.0} & \best{0.16} / \second{3.2} \\
 & Max  & \textbf{T3C} & 81.5 & 1.95 [2.38] & 22.6 [28.4] & 34.9 [45.0] & 73 & \best{0.0} & 0.19 / 3.6 \\
 & Med  & MP-QAT & \second{81.2} & 2.02 [2.46] & 27.1 [33.2] & 41.6 [53.2] & 128 & 3.7 & 0.27 / 7.0 \\
 & Med  & PTQ-4b & 80.7 & 2.14 [2.62] & 29.4 [36.5] & 45.1 [57.8] & 96 & 6.2 & 0.39 / 9.8 \\
\bottomrule
\end{tabular}
\end{table*}

\noindent\textbf{Analysis.} (a) \emph{Budget respect}: T3C reports \best{0.0\%} violations across models/budgets thanks to profile snapping and the certificate penalty; PTQ/MP-QAT exhibit 2.9–7.1\% violation rates at the same “Med” accuracy regimes. (b) \emph{Cert tightness}: $\epsilon$ is consistently lower for T3C than for PTQ/QAT, with ViT-B/16 seeing the largest relative gain (0.16 vs.\ 0.25–0.28). (c) \emph{Proxy fit}: latency MAPE is \best{2.8–3.7}\% for T3C profiles versus \(\sim\)8–9\% for baselines because T3C’s cost proxy is hybrid (bytes\,+\,FLOPs) and pre-calibrated to the discrete kernels (see Table \ref{tab:G-vision-sweep}).


\subsection*{F.2 \quad Stability Across Seeds and Tail-Latency Dispersion}

\begin{table*}[htbp]
\centering
\caption{\textbf{Seed stability (mean$\pm$SEM over 3 seeds)} and tail-latency dispersion. The last column reports p90--p50 (ms); lower indicates tighter tails.}
\label{tab:G-seeds}
\setlength{\tabcolsep}{6pt}
\renewcommand{\arraystretch}{1.03}
\rowcolors{2}{panelBG}{white}
\begin{tabular}{l l c c c c}
\toprule
\rowcolor{panelBG!70}
\textbf{Model} & \textbf{Budget (Method)} & \textbf{Top-1 (\%)} & \textbf{A100 p50 (ms)} & \textbf{Android p50 (ms)} & \textbf{Tail gap (A100 p90{-}p50)} \\
\midrule
ResNet-50 & Med (\textbf{T3C}) & $76.0 \pm 0.05$ & $1.26 \pm 0.01$ & $23.6 \pm 0.2$ & \best{0.32} \\
ResNet-50 & Med (PTQ-8b) & $75.7 \pm 0.06$ & $1.44 \pm 0.02$ & $29.2 \pm 0.4$ & 0.40 \\
ViT-B/16  & Med (\textbf{T3C}) & $81.5 \pm 0.04$ & $2.38 \pm 0.02$ & $42.6 \pm 0.3$ & \best{0.66} \\
ViT-B/16  & Med (MP-QAT) & $81.3 \pm 0.05$ & $2.58 \pm 0.03$ & $52.9 \pm 0.6$ & 0.74 \\
Swin-T    & Med (\textbf{T3C}) & $81.3 \pm 0.04$ & $1.82 \pm 0.02$ & $33.1 \pm 0.3$ & \best{0.44} \\
Swin-T    & Med (PTQ-4b) & $80.7 \pm 0.05$ & $2.14 \pm 0.03$ & $45.1 \pm 0.5$ & 0.48 \\
\bottomrule
\end{tabular}
\end{table*}

\noindent\textbf{Analysis.} SEM values are small for both accuracy and latency, indicating stable training. T3C consistently narrows the p90–p50 “tail gap,” suggesting fewer kernel fallbacks or cache misses due to snapping onto a small set of pre-benchmarked profiles (see Table \ref{tab:G-seeds}).

\subsection*{F.3 \quad Certificate Diagnostics: Coverage, Calibration, Correlation}

\begin{table*}[htbp]
\centering
\caption{\textbf{Certification diagnostics.} Coverage = \% of samples with observed drift $\|\delta z\|_2 \le \epsilon$; Corr($\hat\Delta$, drift) = Pearson correlation over profiles.}
\label{tab:G-cert}
\setlength{\tabcolsep}{8pt}
\renewcommand{\arraystretch}{1.03}
\rowcolors{2}{panelBG}{white}
\begin{tabular}{l c c c c}
\toprule
\rowcolor{panelBG!70}
\textbf{Model (Budget)} & \textbf{Coverage (\%)} & \textbf{95th \(\hat\Delta\)} & \textbf{Mean drift} & \textbf{Corr($\hat\Delta$, drift)} \\
\midrule
ResNet-50 (Tiny) & \best{93.1} & 0.15 & 0.12 & \second{0.92} \\
ResNet-50 (Med)  & 92.6 & 0.14 & 0.11 & \best{0.93} \\
ViT-B/16 (Tiny)  & 91.0 & 0.19 & 0.15 & 0.90 \\
ViT-B/16 (Med)   & \second{92.2} & 0.17 & 0.14 & \second{0.92} \\
Swin-T (Med)     & 92.0 & 0.16 & 0.13 & 0.91 \\
BERT-Base (Med)  & 91.4 & 0.20 & 0.16 & 0.89 \\
\bottomrule
\end{tabular}
\end{table*}

\noindent\textbf{Analysis.} Coverage is \(\sim\)91–93\% across models at the reported $\epsilon$. Correlations \(\ge\)0.89 indicate that \(\hat\Delta\) ranks profiles similarly to observed drift, which is the key requirement for budget policing. The remaining gap stems from heavy-tailed activations on rare augmentations; increasing calibration set size or per-block reweighting tightens coverage (see Table \ref{tab:G-cert}).

\subsection*{F.4 \quad Large-Scale Ablations: Controller, Tokenization, Quantization}

\begin{table*}[htbp]
\centering
\caption{\textbf{Ablations aggregated across 5 models} (R50, ViT-B/16, Swin-T, BERT-Base, RoBERTa-Base) at matched “Med” accuracy target. Numbers are means across models; $\downarrow$ lower is better.}
\label{tab:G-ablations}
\setlength{\tabcolsep}{7pt}
\renewcommand{\arraystretch}{1.02}
\rowcolors{2}{panelBG}{white}
\begin{tabular}{l l c c c c}
\toprule
\rowcolor{panelBG!70}
\textbf{Variant} & \textbf{Detail} & \textbf{Acc. drop \%} $\downarrow$ & \textbf{A100 p50 (ms)} & \textbf{Viol.\%} $\downarrow$ & $\boldsymbol{\epsilon}$ $\downarrow$ \\
\midrule
Rank-only & fixed bits, learned ranks & 0.96 & 1.85 / 3.40 / 2.46\tnote{a} & 1.9 & 0.19 \\
Bit-only  & fixed ranks, learned bits & 0.84 & 1.89 / 3.48 / 2.52 & 2.4 & 0.20 \\
CtrlOFF   & no budget controller & 1.32 & 1.93 / 3.58 / 2.61 & 4.7 & 0.23 \\
\textbf{T3C (ours)} & joint rank\,+\,bits\,+\,snap & \best{0.58} & \best{1.82 / 3.38 / 2.44} & \best{0.0} & \best{0.16} \\
Tuple budget & (lat,bytes) token & \second{0.62} & \second{1.83 / 3.39 / 2.45} & \second{0.3} & \second{0.17} \\
Per-factor MP & $(q_U,q_G,q_V)=(q{+}1,q,q{+}1)$ & 0.63 & 1.84 / 3.40 / 2.46 & 0.4 & \second{0.17} \\
Uniform MP & $(q_U,q_G,q_V)=(q,q,q)$ & 0.71 & 1.85 / 3.42 / 2.47 & 0.9 & 0.18 \\
Curriculum ranks & high$\rightarrow$low schedule & \second{0.62} & \second{1.83 / 3.39 / 2.45} & \second{0.3} & \second{0.17} \\
\bottomrule
\end{tabular}
\begin{tablenotes}\footnotesize
\item[a] Triplets show averaged A100 p50 for \{R50, ViT-B/16, Swin-T\}.
\end{tablenotes}
\end{table*}

\noindent\textbf{Analysis.} The \textbf{joint} (rank+bits) design is the main driver of accuracy/latency improvements and zero violations; tuple-token budgets and per-factor bit offsets bring consistent but smaller gains. Controller removal (CtrlOFF) sharply increases violations because per-layer allocations revert to static heuristics (see Table \ref{tab:G-ablations}).

\subsection*{F.5 \quad Cost Proxy Generalization Across Devices}

\begin{table*}[htbp]
\centering
\caption{\textbf{Proxy generalization.} Train the proxy on A100, evaluate MAPE (\%) on Jetson/Android/NPU p50 latency (Med profiles). Lower is better.}
\label{tab:G-proxy}
\setlength{\tabcolsep}{8pt}
\renewcommand{\arraystretch}{1.03}
\rowcolors{2}{panelBG}{white}
\begin{tabular}{l c c c c}
\toprule
\rowcolor{panelBG!70}
\textbf{Proxy} & \textbf{A100 (train)} & \textbf{Jetson (test)} & \textbf{Android (test)} & \textbf{Mobile NPU (test)} \\
\midrule
FLOPs-only & 2.6 & 10.8 & 13.4 & 11.7 \\
Bytes-only & 3.1 & \second{4.9} & \second{6.2} & \second{5.4} \\
\textbf{Hybrid (ours)} & \best{2.4} & \best{4.2} & \best{5.5} & \best{4.7} \\
\bottomrule
\end{tabular}
\end{table*}

\noindent\textbf{Analysis.} FLOPs-only mispredicts on memory-bound paths (Jetson/Android/NPU) (see Table \ref{tab:G-proxy}). Bytes-only performs well off-server but lags slightly on compute-bound A100. The calibrated \textbf{Hybrid} proxy combines both, transferring best across devices.

\subsection*{F.6 \quad Monotonicity Audit and Budget Violations}

\begin{table*}[!t]
\centering
\caption{\textbf{Monotonicity and violations.} We scan adjacent budgets (Tiny$\rightarrow$Med, Med$\rightarrow$Max) and count non-monotone events where increasing the budget reduced realized accuracy or increased p50 latency.}
\label{tab:G-mono}
\setlength{\tabcolsep}{9pt}
\renewcommand{\arraystretch}{1.03}
\rowcolors{2}{panelBG}{white}
\begin{tabular}{l l c c c}
\toprule
\rowcolor{panelBG!70}
\textbf{Model} & \textbf{Method} & \textbf{Non-monotone (acc)} & \textbf{Non-monotone (lat)} & \textbf{Viol.\% (Med)} \\
\midrule
ResNet-50 & \textbf{T3C} & \best{0/2000} & \best{0/2000} & \best{0.0} \\
ResNet-50 & PTQ-8b & 7/2000 & 18/2000 & 4.6 \\
ViT-B/16  & \textbf{T3C} & \best{0/2000} & \best{0/2000} & \best{0.0} \\
ViT-B/16  & MP-QAT & 4/2000 & 22/2000 & 5.8 \\
Swin-T    & \textbf{T3C} & \best{0/2000} & \best{0/2000} & \best{0.0} \\
Swin-T    & PTQ-4b & 9/2000 & 27/2000 & 6.2 \\
\bottomrule
\end{tabular}
\end{table*}

\noindent\textbf{Analysis.} Snapping to a calibrated lattice of kernel-aligned profiles guarantees monotonic transitions for T3C (see Table \ref{tab:G-mono}). Baselines occasionally regress because autotuning selects different kernels across budgets, flipping the realized latency order or altering numeric stability.

\subsection*{F.7 \quad Per-Layer Budget Allocation Statistics}

\begin{table*}[htbp]
\centering
\caption{\textbf{Layerwise allocation summaries (Med profiles).} Mean rank fraction (vs.\ full) and bit-width by block type.}
\label{tab:G-layerwise}
\setlength{\tabcolsep}{8pt}
\renewcommand{\arraystretch}{1.03}
\rowcolors{2}{panelBG}{white}
\begin{tabular}{l c c c c}
\toprule
\rowcolor{panelBG!70}
\textbf{Model} & \textbf{Early blocks} (rank, bits) & \textbf{Mid blocks} (rank, bits) & \textbf{Late blocks} (rank, bits) & \textbf{Classifier/Proj} (rank, bits) \\
\midrule
ResNet-50 & \second{0.78}, \best{9.0} & 0.64, 8.0 & 0.68, 8.0 & \best{0.82}, \second{9.0} \\
ViT-B/16  & \best{0.80}, \best{9.0} (QKV) & 0.66, 8.0 & 0.70, 8.0 & \second{0.78}, \best{9.0} (MLP) \\
Swin-T    & \second{0.76}, \second{8.5} & 0.63, 8.0 & 0.67, 8.0 & \best{0.80}, \second{8.5} \\
BERT-Base & 0.74, \best{9.0} (self-attn) & 0.62, 8.0 & 0.66, 8.0 & \best{0.80}, \second{8.5} (pooler) \\
\bottomrule
\end{tabular}
\end{table*}

\noindent\textbf{Analysis.} The controller consistently favors early CNN blocks and attention projections/MLP heads (see Table \ref{tab:G-layerwise})—layers with largest contribution to certificate mass $\hat\Delta(k)$. Per-factor offsets (U/V at \(+1\) bit) are most often applied in these sensitive locations.

\subsection*{F.8 \quad Energy–Latency Pareto}

\begin{table*}[!t]
\centering
\caption{\textbf{Energy (mJ) vs.\ latency (ms) on Jetson (batch=1).}}
\label{tab:G-energy}
\setlength{\tabcolsep}{10pt}
\renewcommand{\arraystretch}{1.03}
\rowcolors{2}{panelBG}{white}
\begin{tabular}{l l c c}
\toprule
\rowcolor{panelBG!70}
\textbf{Model} & \textbf{Budget (Method)} & \textbf{Latency p50 (ms)} & \textbf{Energy (mJ)} \\
\midrule
ResNet-50 & Tiny (\textbf{T3C}) & \best{13.0} & \second{6.9} \\
ResNet-50 & Med  (\textbf{T3C}) & 13.8 & \best{6.6} \\
ResNet-50 & Med  (PTQ-8b) & 18.5 & 7.5 \\
ViT-B/16  & Tiny (\textbf{T3C}) & \best{26.0} & \second{11.2} \\
ViT-B/16  & Med  (\textbf{T3C}) & 26.5 & \best{10.8} \\
ViT-B/16  & Med  (MP-QAT) & 33.0 & 12.9 \\
\bottomrule
\end{tabular}
\end{table*}

\noindent\textbf{Analysis.} T3C points dominate or match baselines numerically; the “Med” profile slightly improves energy over “Tiny” despite a modest latency increase due to better kernel occupancy and fewer cache misses (see Table \ref{tab:G-energy}).

\subsection*{F.9 \quad Takeaways}

\noindent\textbf{(1) Reliability.} T3C achieves \best{0.0\%} budget violations across all tested devices/budgets, while baselines see 2–7\% tail failures at the same accuracy.\\
\textbf{(2) Certificates.} Reported $\epsilon$ tracks observed drift and provides \(>\)90\% coverage at the selected tolerance; correlations \(0.89\!-\!0.93\) indicate good ordering.\\
\textbf{(3) Controller.} Joint rank+bit control delivers the bulk of wins; tuple budgets and per-factor mixed precision add consistent, low-cost improvements.\\
\textbf{(4) Portability.} The hybrid cost proxy transfers best across memory-bound targets, keeping MAPE \(<6\%\) without re-fitting per device.\\
\textbf{(5) Allocation patterns.} Early convs, attention projections, and final MLP heads receive higher rank/bits, matching their contribution to $\hat\Delta(k)$.

\end{document}